\documentclass[10pt,twocolumn,letterpaper]{article}

\usepackage{iccv}              % To produce the CAMERA-READY version
\usepackage{amsmath, amssymb, graphicx}
\definecolor{iccvblue}{rgb}{0.21,0.49,0.74}
\usepackage[pagebackref,breaklinks,colorlinks,allcolors=iccvblue]{hyperref}

\usepackage{times}

\usepackage{epsfig}
\usepackage{graphicx}
\usepackage{amsmath}
\usepackage{amssymb}
\usepackage{arydshln}
\usepackage{enumitem}
\usepackage{subcaption}

\usepackage{listings}

\usepackage{microtype}
\usepackage{booktabs} % for professional tables
\usepackage{multirow}
\usepackage{tikz}
\usepackage{dblfloatfix}
\usepackage{pgfplots}
\pgfplotsset{compat=1.16}
\usepgfplotslibrary{groupplots}
\usetikzlibrary{patterns}

\usepackage{tcolorbox}

\usetikzlibrary{arrows}
\usepackage{tkz-kiviat}
\usepackage{xfp}

\definecolor{6ec2b5}{HTML}{6ec2b5}
\colorlet{Dark}{6ec2b5}

\usepackage{xcolor}
\definecolor{citecolor}{HTML}{0071bc}
\definecolor{color_ao}{gray}{0.5}
\definecolor{color_our}{HTML}{78debc}
\definecolor{color_pre}{rgb}{0.52,0.59,0.69}
\definecolor{Gray}{gray}{0.9}
\definecolor{LighterGray}{gray}{0.93}
\definecolor{LightGrayForTableRule}{gray}{0.92}
\definecolor{DarkGray}{gray}{0.5}
\definecolor{Black}{rgb}{0.0, 0.0, 0.0}
\definecolor{NiceBlue}{rgb}{0.11764705882352941, 0.5647058823529412, 1.0}
\definecolor{NiceBlue2}{HTML}{3399FF}
\definecolor{NiceGreen}{HTML}{008080}
\usepackage{diagbox}
\definecolor{gray}{HTML}{b0aeae}
\definecolor{diffgray}{HTML}{878787}
\definecolor{NiceGray}{HTML}{696969}
\definecolor{applegreen}{rgb}{0.55, 0.71, 0.0}

\definecolor{demphcolor}{RGB}{144,144,144}

\hypersetup{
    colorlinks,
    linkcolor={red},
    citecolor={Dark}
}

\usepackage{mathtools}
\usepackage{amsthm}

\usepackage{xcolor,colortbl}
\definecolor{Gray}{gray}{0.90}
\newcolumntype{g}{>{\columncolor{Gray}}c}
\definecolor{ffe1da}{RGB}{255,225,218}
\definecolor{E6F5F0}{HTML}{E6F5F0}
\definecolor{darkF7E0D5}{RGB}{209,154,128}
\definecolor{F5F9FF}{HTML}{edf6ff}
\colorlet{Light}{E6F5F0} % {E6F5F0}
\newcommand{\CC}[1]{\cellcolor{Light}}

\def\ourapproach{\textsc{Agentic-DRS}\xspace}
\def\ourbenchmark{\textsc{DRS-Bench}\xspace}
\def\ourdataset{\text{IDD}\xspace}

\usepackage{amssymb}% http://ctan.org/pkg/amssymb
\usepackage{pifont}% http://ctan.org/pkg/pifont

\usepackage{graphbox}
\usepackage{booktabs}       % professional-quality tables
\usepackage{amsfonts}       % blackboard math symbols
\usepackage{nicefrac}       % compact symbols for 1/2, etc.
\usepackage{microtype}      % microtypography
\usepackage{comment}
\usepackage{amsmath,amssymb} % define this before the line numbering.
\usepackage{caption}
\usepackage{float}
\usepackage{algorithm}
\usepackage[noend]{algpseudocode}

\title{\vspace{0pt}Agentic Design Review System}

\author{Sayan Nag, K J Joseph, Koustava Goswami, Vlad I Morariu, Balaji Vasan Srinivasan\\
Adobe Research
}

\begin{document}
\maketitle
\begin{abstract}
Evaluating graphic designs involves
assessing it from multiple facets like alignment, composition, aesthetics and color choices. Evaluating designs in a holistic way involves aggregating feedback from individual expert reviewers. Towards this, we propose an Agentic Design Review System (Agentic-DRS), where multiple agents collaboratively analyze a design, orchestrated by a meta-agent. A novel in-context exemplar selection approach based on graph matching and a unique prompt expansion method plays central role towards making each agent design aware.
Towards evaluating this framework, we propose \ourbenchmark benchmark. Thorough experimental evaluation against state-of-the-art baselines adapted to the problem setup, backed-up with critical ablation experiments brings out the efficacy of Agentic-DRS in evaluating graphic designs and generating actionable feedback. We hope that this work will attract attention to this pragmatic, yet under-explored research direction \footnote{Web page: \url{https://sayannag.github.io/AgenticDRS/}}.

    % Design evaluation is a complex and multi-faceted process requiring structured assessments across multiple attributes, such as alignment, aesthetics, styles, grouping, etc. Traditional heuristics-based evaluation methods often lack adaptability and structured reasoning, leading to inconsistencies and inefficiencies. In this work, we introduce an \textbf{Agentic}-\textbf{D}esign \textbf{R}eview \textbf{S}ystem (DRS) for automated design evaluation, wherein a multi-agent system collaboratively analyzes design artifacts by leveraging structured reasoning and graph-based retrieval mechanisms. Our framework consists of a meta-agent orchestrating a set of static and dynamic design reviewers, each specializing in different evaluation criteria. Furthermore, we incorporate an interpretable graph-based representation of designs leveraging semantic and spatial intra-design element relationships, enabling similarity-based in-context example selection via node-matching and edge-matching algorithms. Experiments demonstrate improved evaluation consistency and adaptability over heuristic and single-agent approaches by 34.0\% and 12.6\% respectively, highlighting the potential of Agentic framework in scalable, interpretable design assessment.
\end{abstract}

\section{Introduction}
Graphic designs like flyers, posters, invitation-cards, \etc are harmonious compositions of images, text, shapes and their colors, nicely laid-out aesthetically, to convey the meaning intended by their designer. They have become ubiquitous in our daily lives from the brochure of your new car to the birthday invitation of your toddler. 

% With the availability of do-it-yourself design tools like Canva \cite{canva}, Adobe Express \cite{express} and Microsoft Designer \cite{designer}, amateurs and novice designers are empowered to create professional designs.
% With the proliferation in the use of such designs in social media platforms, these tools are getting increasingly popular.
With the availability of do-it-yourself design tools, amateurs and novice designers are empowered to create professional designs.
With the proliferation in the use of such designs in social media platforms, these tools are getting increasingly popular.
% Such tools are getting increasingly popular paired with the demand for such designs for social-media usage.
% ; for instance, Adobe Express records $86\%$ year-over-year increase in the cumulative number of creations made with its platform \cite{adobe}. \balaji{This is an Adobe stat - please remove it. Has a lot of implications when it comes from us. This might be a good motivation for an invention disclosure, not so much for a paper. }
Novice designers lack deep understanding of design principles like balance, emphasis, unity, white-space usage and so on, which would have a profound impact of their final generation. 
A tool that can \textit{automatically analyze a design and provide actionable feedback} would be of immense value for such designers.  

With the recent advancements in Multi-modal LLMs \cite{liu2023visual,zhang2024internlm} and Diffusion Models \cite{rombach2022high}, researchers have introduced novel approaches \cite{jia2023cole, inoue2024opencole} for generating graphic designs from textual prompts. This is a significant advancement from the earlier efforts \cite{inoue2023towards, levi2023dlt, luo2024layoutllm, lin2023layoutprompter} which generate just the layout (positioning information of elements), to generating the entire design, with the content filled in. As these technologies mature, they would truly enable Human-AI co-creation for design generation. A yard-stick to measure the progress of design generation methods would be a design-evaluator that \textit{introspects a design across multiple dimensions} such as typographic quality, color consistency and semantic coherence. 
\begin{figure}[t!]
    \centering
    \includegraphics[width=0.48\textwidth]{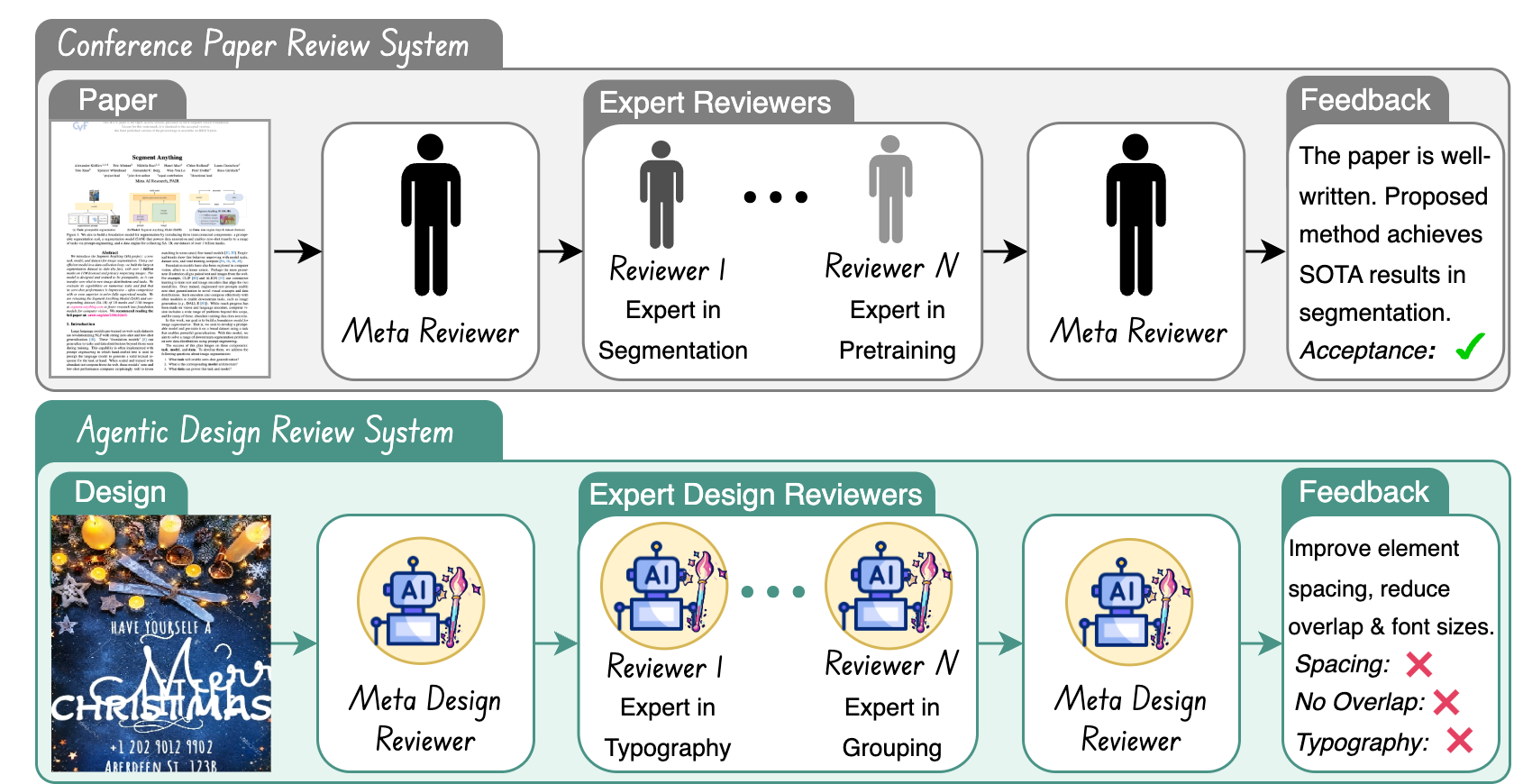}
    \caption{Evaluating a graphic design involves assessing it across multiple dimensions like visual coherence, semantic grouping, typographic clarity and so on. Inspired by peer-review system in conferences, we propose to build the \textit{first Agentic framework for design evaluation and feedback generation}. We propose a novel approach to infuse design knowledge into agents, and allow them to collaboratively review and access the input design in \cref{sec:method}.}
    \label{fig:teaser}
\end{figure}

Judging whether a design is good or bad is hard and subjective as characteristics of designs are inherently \textit{tacit}\footnote{Dictionary meaning: understood or implied without being stated.} \cite{son2024demystifying}. Tacit principles like color pairing for a specific demography, is hard to objectively define and quantify making design evaluation extremely challenging. Heuristic approaches \cite{odonovan, ngo2000mathematical, ngo2001another, harrington2004aesthetic, bauerly2006computational, scholgens2016aesthetic, zen2014towards} try to mathematically quantify design characteristics like alignment, overlap and whitespace to give a score, but fail to see the overall global harmony of the designs. Learning based approaches \cite{zhao2018characterizes,sawada2024visual, dou2019webthetics, wan2021novel} treats the design evaluation as either a regression or classification task. Presence of a diverse enough dataset is a prerequisite for its performance. Recently, evaluations form GPT-4o \cite{gpt4o} has found to correlate well with human judgment of graphic designs \cite{gde}. Building on this trend, we propose a holistic evaluation suite to access the quality of graphic designs, and generate actionable feedback for the designers.

By virtue of the huge amount of training data, and the learning paradigm, Multimodal LLMs (MLLMs) such as  GPT-4o \cite{gpt4o} posses novice-level awareness of the characteristics of a good graphic design. Our first contribution is to enhance the design awareness of these models using a novel graph matching based in-context exemplar selection approach (\cref{sec:grad}) and a structured description based prompt expansion strategy (\cref{sec:sdd}). 
A single instance of such a design-aware MLLM would not be able to assess a design across the variety of dimensions on which graphic designs should be evaluated. Towards this end, we propose the \textit{first Agentic framework for design evaluation and feedback generation}. Each agent will be specializing in a specific aspect of evaluation like color harmony, typographic quality, alignment consistency and so on. Given an input design, a meta-agent spawns-off these agents as necessary (could even dynamically control what aspect the agent should evaluate a design on, referred to as dynamic agents in \cref{sec:method}), and collates the independent feedback after the review process to generate scores and actionable feedback.

Towards evaluating our approach, we introduce \ourbenchmark, a holistic benchmark suite containing $15$ design attribute definitions, $4$ datasets, new evaluation metrics and strong baselines. Our experimental analysis showcases the effectiveness of using design-aware MLLMs in an Agentic framework for design evaluation and feedback generation. 

% We evaluate our proposed approach across multiple baselines and we clearly outperform them (\joseph{will update, this is a placeholder}).
%Further, we introduce a taxonomy of design attributes across which designs should be evaluated by gathering inputs from $20$ expert designers.
% \balaji{If this is a work-in-progress, then fine. I would guard against revealing too much on the design attribute data - that is an internal data. We have to see how much we can reveal about that. Also, since there is an active interest in the product here, it might raise questions about business risk.}

\vspace{3pt}
\noindent Our key contributions are summarized below:
\begin{itemize}[leftmargin=*]
\setlength\itemsep{-0.05em}
    \item We introduce the first Agentic evaluation framework that can score designs and generate actionable feedback. 
    \item We enhance the design awareness of Multi-modal LLMs with a novel graph matching based exemplar selection approach and  structured description based prompt expansion.
    \item We introduce \ourbenchmark, a holistic framework for assessing design evaluation quality and feedback generation.
    % \item We rely on a taxonomy of attributes identified by $20$ expert designers as important to evaluating designs. \balaji{I would obfuscate this a bit.}
    \item Through rigorous experimentation, we bring out the efficacy of our proposed approach, clearly out-performing the state-of-the-art baselines adapted to the task.
\end{itemize}

% \joseph{work-in-progress}
% \vlad{I know this is in-progress, but just so we don't forget -- we should ensure to highlight current shortcomings of earlier design evaluation methods before introducing ours}

\section{Related Work}

% \koustava{Work In Progress}

% \paragraph{Design Heuristics based Approaches.}

% \paragraph{Data Driven Approaches.}
\noindent \textbf{Design Evaluation.} Design evaluation has been widely studied across domains and development stages. Early heuristic-based methods \cite{DBLP:journals/ijmms/BauerlyL06,DBLP:conf/doceng/HarringtonNJRT04,ngo2000mathematical,odonovan,DBLP:conf/rcis/ZenV14} struggled to capture semantic design attributes. Machine learning models \cite{DBLP:journals/ijmms/DouZSH19} improved evaluation but lacked comparative scoring mechanisms, leading to siamese-based approaches \cite{DBLP:journals/tog/ZhaoCL18, goyal2024design}. \citet{DBLP:journals/tvcg/KongJSGCLLZ23} trained a model for aesthetic assessment, while \citet{DBLP:conf/siggraph/TabataYMY19} perturbed layouts to generate poor designs for scoring. Recent MLLM-based methods focus on design generation \cite{DBLP:journals/corr/abs-2311-16974} or assess limited attributes \cite{DBLP:journals/corr/abs-2311-16974, gde} but lack comprehensive, actionable feedback grounded in a comprehensive set of design attributes, an aspect we address.

\noindent \textbf{MLLM Agents.} Agentic workflows using large models are widely adopted for reasoning-based tasks. While early visual agents were fine-tuned for specific tasks \cite{DBLP:conf/cvpr/ShridharTGBHMZF20}, recent MLLMs enable broader agentic workflows \cite{DBLP:journals/corr/abs-2402-15116}. These models integrate multi-modal encoders, allowing agents to tackle diverse tasks, including games \cite{DBLP:conf/rss/BrohanBCCDFGHHH23,DBLP:conf/eccv/YangDLLWTJKZZL24,DBLP:conf/icml/DriessXSLCIWTVY23}, design \cite{DBLP:journals/corr/abs-2403-03163,DBLP:journals/corr/abs-2403-09029}, human-computer interaction \cite{DBLP:conf/icml/ZhengGK0024,DBLP:conf/iclr/ZhouX0ZLSCOBF0N24,DBLP:conf/eccv/KapoorBRKKAS24,DBLP:journals/corr/abs-2312-13771}, and audio understanding \cite{DBLP:journals/corr/abs-2310-12404,DBLP:conf/emnlp/YuSLH00Z023,DBLP:conf/aaai/HuangLYSCYWHHLR24,DBLP:journals/corr/abs-2307-14335}. These agents use structured planning and reasoning for complex problem-solving. However, to the best of our knowledge, no prior work has explored a multi-agent, multi-modal agentic workflow for Design Evaluation. We take the first step in this direction.

\noindent \textbf{In-context Learning in MLLMs.} In-context learning (ICL), where example inputs and outputs are included in prompts, is widely used in NLP to guide models toward relevant information for downstream tasks \cite{DBLP:conf/emnlp/Dong0DZMLXX0C0S24,DBLP:journals/corr/abs-2212-06713,DBLP:conf/acl/Honovich0BL23,DBLP:conf/acl/MinLHZ22,DBLP:conf/acl/MosbachPRKE23}. ICL has been extended to multi-modal applications \cite{DBLP:journals/corr/abs-2309-04790,DBLP:conf/cvpr/ZhangLYWLL0W24,DBLP:conf/iclr/ZhaoCSMA0LWHC24}, improving MLLM reasoning primarily for text-based tasks \cite{DBLP:conf/nips/AlayracDLMBHLMM22,DBLP:conf/nips/JiangJXYDY00Z24,DBLP:conf/nips/LaurenconSTBSLW23,DBLP:journals/corr/abs-2306-05425,DBLP:journals/corr/abs-2407-07895}. \citet{DBLP:journals/corr/abs-2411-11909} demonstrated that visual context enhances ICL reasoning but within a fine-tuning framework. Training-free ICL sample selection has also been explored \cite{ram2023context, ferber2024context}, though limited to text \cite{ram2023context} or global features \cite{ferber2024context}. In contrast, given our focus on design evaluation, we introduce a graph-based retrieval mechanism leveraging local features for in-context exemplar design retrieval while preserving fine-grained structural and semantic relationships among the various design elements.

\section{Methodology} \label{sec:method}

\begin{figure*}[t!]
    \centering
    \includegraphics[width=\textwidth]{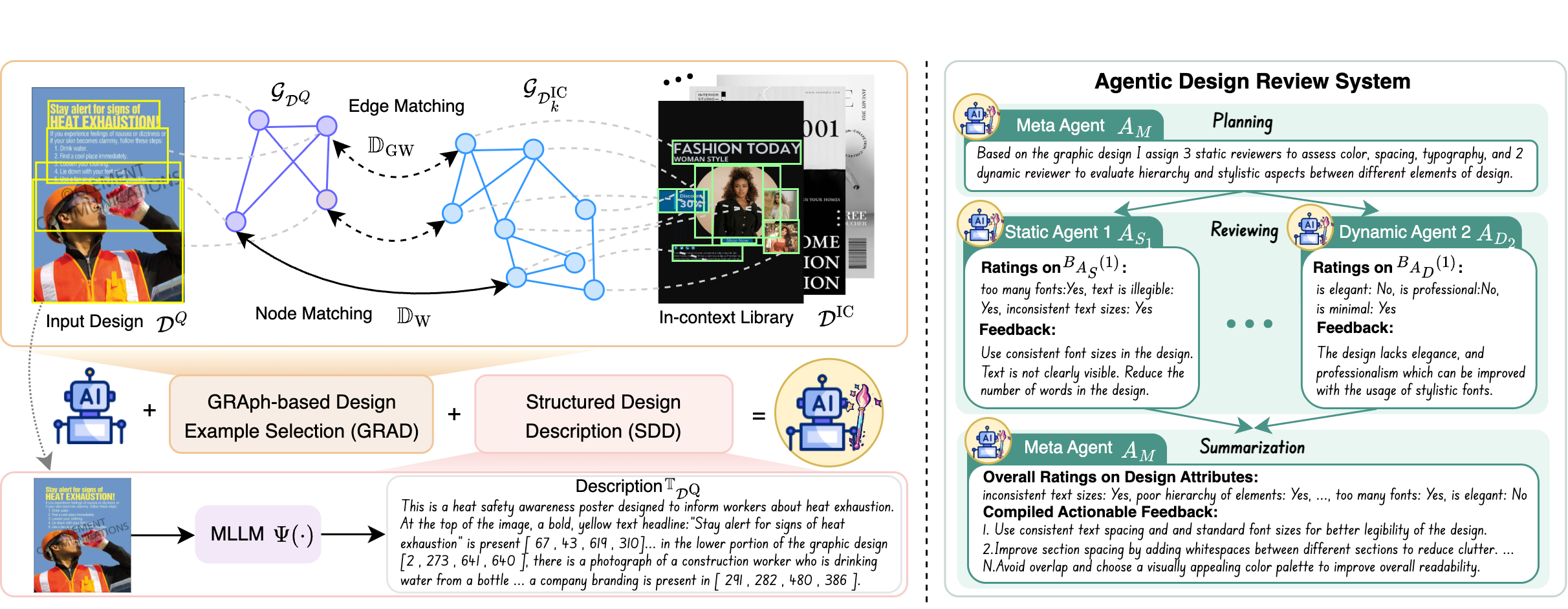}
    \vspace{-15pt}
    \caption{\textbf{Overview of the proposed design evaluation pipeline.} To systematically evaluate designs, GRAD constructs a graph representation of design elements using edge and node matching techniques, enabling structured retrieval of in-context design examples from a curated library (Sec \ref{sec:grad}). Structured Design Description (SDD) module generates design descriptions (Sec \ref{sec:sdd}) to anchor the responses of each agent. The selected $K$ designs, query design along with its description inform the review process, where a meta agent coordinates static and dynamic agents to assess disparate design attributes - part of the Agentic Design Review System. Static agents focus on a fixed set of attributes (e.g., typography), whereas dynamic agents evaluate the attributes which are contextualized to specific designs (e.g., stylistic qualities). The meta agent consolidates these insights into a final rating and provides actionable feedback for design improvements (Sec \ref{sec:adrs}).}
    \label{fig:mainfig}
    \vspace{-10pt}
\end{figure*}

% This multi-agent framework enhances automated design evaluation by leveraging structured reasoning and comparative analysis.

% \subsection{Adapting MLLMs for Designs}

Evaluating designs and generating actionable feedback are non-trivial and inherently complex tasks, requiring a nuanced understanding of aesthetics, functionality, user expectations, and intricacies of designs. The seminal work from Haraguchi \etal \cite{gde}, shows how  MLLMs such as GPT-4o are able to evaluate designs 
correlating with human ratings. This is an important first step towards evaluating designs in a systematic manner. 
% have made an important first step towards a unified framework for evaluating three key design aspects (alignment, overlap, and white space) in a systematic manner .
Building on this, we further enhance the design capabilities of these MLLMs by making them more design-aware, in a \textit{training-free} way, by introducing a novel graph-based in-context design selection approach (Sec \ref{sec:grad}) and anchoring MLLMs for designs with visually grounded structured descriptions (Sec \ref{sec:sdd}). Given a query design to be evaluated, these key innovations assist in dynamically retrieving semantically and structurally relevant design examples from a design data library, thereby improving contextual understanding and enhancing the accuracy and robustness of design evaluation.

Scoring and generating feedback for a design should account for multiple aspects like visual appeal, semantic coherence, typographic quality, \etc.  Towards generating a holistic feedback that accounts for these multiple perspectives, we draw inspiration from how humans review their creative generations. The peer-review process in conferences solves a similar task by having a group of expert reviewers with diverse specializations (or expertise) analyzing the same subject from multiple perspectives, and meta-reviewers aggregating their opinions.
% Furthermore, human decision review systems, e.g., as used in peer-reviewed conferences, operate with a group of expert reviewers with diverse specializations (or expertise) analyzing the same subject from multiple perspectives and considering multiple attributes. 
This naturally inspires us to propose an Agentic Design Review System (as demonstrated in Fig \ref{fig:teaser}), which to the best of our knowledge is a first effort to formalize such a collaborative review system for graphic design evaluation in Sec \ref{sec:adrs}. By mirroring the principles of expert-driven decision-making, \ourapproach not only improves the robustness of design evaluations but also fosters a more explainable and structured critique process. This allows for nuanced feedback that balances subjective creativity with objective design principles, enabling designers to receive context-aware, multi-faceted insights that drive meaningful improvements in their work.

\subsection{Graph-based Design Exemplar Selection}\label{sec:grad}
Conventional in-context example selection approaches often rely on global CLIP feature-based similarity, which may overlook finer design relationships such as spatial composition, alignment, grouping information, reading order, and contextual interplay among elements. Our \textbf{GRA}ph-based \textbf{D}esign exemplar selection method (GRAD) encodes \textit{semantic}, \textit{spatial}, and \textit{structural} relationships by constructing localized representations of the graphic designs by offering certain advantages: \textit{(i)} Preservation of structural and semantic relationships: unlike global feature matching , our graph representation captures relative positions and proximity of design elements. \textit{(ii)} Adaptability to open-world designs: by computing similarities based on both semantic and spatial embeddings, the selection method generalizes across diverse design styles. To facilitate structured comparison and retrieval of reference designs, we define a query design (to be evaluated) $\mathcal{D}^{Q}$, and designs present in the In-context (IC) data library $\mathcal{D}^{\text{IC}} \in \mathcal{D}$ in the form of graphs $\mathcal{G}_{\mathcal{D}^{j}} = (\mathbb{V}^{j}, \mathbb{E}^{j}), \; j \in \{Q, \text{IC}\}$, where $\mathbb{V}$ represents the vertices, and $\mathbb{E}$ denotes the edges of the graph, the superset of designs is $\mathcal{D}$. Graph creation is done in two ways:

\noindent \textbf{(i) Presence of bounding boxes:} in the presence of layout information metadata for designs, the bounding box (i.e., coordinate) information $\mathcal{BB}^{j}$ can be obtained for all the elements in a graphic design $\mathcal{D}^{j}$. The design components $V^{j}$ can be typically obtained by cropping the design with the coordinate information $\mathcal{BB}^{j}$: $V^{j}(i) = \text{Crop}(D^{j}, \mathcal{BB}^{j}(i)), \; \forall i \in {1, ..., N_{V^{j}}}$, $N_{V^{j}}$ being number of design elements in $\mathcal{D}^{j}$ ($j \in \{Q, \text{IC}\}$), $\text{Crop}(\cdot)$ being the cropping operation. Each such textual and visual elements are subsequently passed through CLIP to obtain respective set of embeddings:
\begin{equation}
    \phi(V^{j})(i) = {\text{CLIP}(V^{j}(i))}, \; \forall i \in {1, ..., N_{V^{j}}}
\end{equation}
These embeddings of the corresponding design elements form the vertices $\mathbb{V} = \{\phi(V^{j})(i)\}_{i=1}^{N_V}$ of the graph. The edges are composed of spatial and semantic distances, where spatial distance $d_{\text{spatial}}(u,v)$ between two elements (i.e., nodes of a graph) $u$ and $v$ is computed based on the normalized L$_{2}$ norm between the bounding box centroids $\mathcal{BB}^{j}(u)$ and $\mathcal{BB}^{j}(v)$ and the semantic distance $d_{\text{semantic}}(u,v)$ is computed as cosine distance between the respective embeddings $\mathbb{V}(u)$ and $\mathbb{V}(v)$. The combined edge weight is given as:
\begin{equation}\label{eq:edge_weight}
    d^{j}(u,v) = d_{\text{spatial}}^{j}(u,v) + d_{\text{semantic}}^{j}(u,v)
\end{equation}

\noindent \textbf{(ii) Absence of bounding boxes:} in the absence of layout information metadata for designs (for datasets without bounding-box details),  we pass the design through a CLIP-ViT encoder and obtain the patch-level features from the last layers of corresponding visual transformer encoders. We subsequently use these encoded local-feature vectors to construct the graph where the vertices $\mathbb{V}$ contain the embedded features of patches and the edges $\mathbb{E}$ comprise of the semantic (cosine) distance between the vertices (patch-level embeddings). Since we do not have a clear structural information present (i.e., absence of boxes) we stick to using semantic relationships only.

Assuming a input design where inherently related elements are not grouped together, our approach GRAD will look for designs in the in-context library not only whose content is semantically close to the design under evaluation (via node matching) but also whose related elements are not clearly grouped (via edge matching). Therefore, GRAD enables structure-aware retrieval, improving the quality of selected in-context examples. This is achieved by the suitable combination of Wasserstein (node matching) \cite{peyre2019ot, cuturi2013sinkhorn, benamou2015iterative, luise2018differential} and Gromov-Wasserstein (edge matching) distances as described below. Given a query design $\mathcal{D}^{Q}$, we select the most relevant in-context examples $\{\mathcal{D}_k^{\text{IC}}\}_{k=1}^{K}$ from a dataset $\mathcal{D}^{\text{IC}}$ using Wasserstein and Gromov-Wasserstein distances.

\noindent \textbf{Wasserstein Distance (WD)} calculates the pairwise distances between the node embedding sets $\mathbb{X}^{Q}$ and $\mathbb{Y}^{k}$ of the respective graphs $\mathcal{G}_{\mathcal{D}^{Q}}$ and $\mathcal{G}_{\mathcal{D}^{\text{IC}}_{k}}$. Considering two discrete distributions, $\psi^{Q} \in \mathbf{P}(\mathbb{X}^{Q})$ and $\psi^{k} \in \mathbf{P}(\mathbb{Y}^{k})$, where $\psi^{Q} = \sum_{i=1}^n z^{Q}(i) \delta_{x^{Q}}(i)$ and $\psi^{k} = \sum_{j=1}^m z^{k}(j) \delta_{y^{k}}(j)$; $\sum_{i}z^{Q}(i) = 1 = \sum_{j}z^{k}(j)$ where $z^{Q}$ and $z^{k}$ are the respective weight vectors for the probability distributions $\psi^{Q}$ and $\psi^{k}$; and $\delta_{x^{Q}}$ is the Dirac-delta function centered on support point $x^{Q}$ in the embedding space. The WD distance between $\psi^{Q}$ and $\psi^{k}$ is defined as: 
\begin{equation}\label{eq:wd}
	\mathbb{D}_{\text{W}}(\mathcal{G}_{\mathcal{D}^{Q}},\mathcal{G}_{\mathcal{D}^{\text{IC}}_{k}}) \triangleq \min_{\mathbf{\Phi}\in \mathbf{\Theta}(z^{Q},z^{k})}\sum_{i,j} \mathbf{\Phi}_{ij} \cdot c(x_i,y_j)
\end{equation}

\noindent where $\mathbf{\Theta}(z^{Q},z^{k}) = \{ \mathbf{\Phi} \in \mathbb{R}_+^{n\times m} | \mathbf{\Phi}\mathbf{1}_m=z^{Q}, \mathbf{\Phi}^\top\mathbf{1}_n=z^{k} \}$, $c(x^{Q}(i),y^{k}(j))$ is cosine distance similarity metric, and $\mathbf{\Phi}$ is the transport plan, interpreting the amount of mass shifted from distributions $\psi^{Q}(i)$ to $\psi^{k}(j)$. An exact solution to the above expression leads to a sparse representation of transport plan $\mathbf{\Phi}$ with at most $(2 \cdot \text{max}(m, n) - 1)$ non-zero elements, ensuring an explainable and robust retrieval \cite{degoes}. A detailed set of steps is provided in Supp Algorithm 1.

\noindent \textbf{Gromov-Wasserstein Distance (GWD)} helps in matching the edges of the graphs and preserves graph topology by computing distances between pairs of nodes thereby ensuring inter-graph structural alignment \cite{peyregromov, alvarez2018gromov}. In the same discrete graph matching setting, GWD can be mathematically represented:
\begin{equation}\label{eq:gwd}
\scriptsize
    \mathbb{D}_\mathrm{GW}(\mathcal{G}_{\mathcal{D}^{Q}},\mathcal{G}_{\mathcal{D}^{\text{IC}}_{k}})
    \triangleq \min_{\hat{\mathbf{\Phi}}\in \mathbf{\Theta}(z^{Q},z^{k})}
\sum_{i,i^{\prime},j,j^{\prime}} 
    \hat{\mathbf{\Phi}}_{ij} \hat{\mathbf{\Phi}}_{i^{\prime}j^{\prime}} 
    \mathcal{L} \left( x^{Q}_i,y^{k}_j,x^{Q}_{i^{\prime}},y^{k}_{j^{\prime}} \right)
\end{equation}

\noindent where inter-graph structural similarity between two node pairs $(x^{Q}_i,x^{Q}_{i^{\prime}})$ and $(y^{k}_j,y^{k}_{j^{\prime}})$ is represented as $\mathcal{L} \left( x^{Q}_i,y^{k}_j,x^{Q}_{i^{\prime}},y^{k}_{j^{\prime}} \right) =\|d^{Q}(x^{Q}_i,x^{Q}_{i^{\prime}}) - d^{k}(y^{k}_j,y^{k}_{j^{\prime}})\|$, $d^{Q}$  being the edge weight between a node pair in graph $\mathcal{G}_{\mathcal{D}^{Q}}$ (see Equation \ref{eq:edge_weight}).  Transport plan $\hat{\mathbf{\Phi}}$ is periodically updated to align the edges in different graphs belonging to disparate designs. A detailed set of steps is provided in Supp Algo 2.

\begin{algorithm}[t!]
\footnotesize
\caption{Overall Pipeline}
\label{alg:algorithm_ADRS}
\begin{algorithmic}[1]
    \Require{Query Design: $\mathcal{D}^{Q}$, In-context Designs: $\mathcal{D}^{\text{IC}}$, In-context samples to be retrieved: $K$, Static Attribute Buckets: $B_{A_S}$, Meta Agent: $A_M$, Static Agents: $A_S$, Number of Static Agents: $N_S$, World Attributes: $\mathcal{W}$, Bounding Boxes: $\mathcal{BB}$, CLIP Encoder: $\text{CLIP}(\cdot)$ Graph Construction Module: $\Omega(\cdot)$, Design Description module: $\Psi(\cdot)$, Wasserstein Distance func.: $\mathbb{D}_{\text{W}}(\cdot)$, Gromov-Wasserstein Distance function: $\mathbb{D}_{\text{GW}}(\cdot)$, Balancing factors: $\alpha$, $\lambda$, Index Sorting function: \texttt{\textbf{Argsort}}($\cdot$), List Element Insertion function: \texttt{\textbf{Insert}}($\cdot$).}
    \Ensure{Dynamic Attribute Buckets: $B_{A_D}$, Dynamic Agents: $A_D$, Number of Dynamic Agents: $N_D$, Feedback: $\mathcal{F}$, Attribute Rating: $\mathcal{R}$, Query Graph: $\mathcal{G}_{\mathcal{D}^{Q}}$, $k^{\text{th}}$ in-context Graph: $\mathcal{G}_{\mathcal{D}^{\text{IC}}_{k}}$, Graph Dissimilarity Scores List: $\mathbb{S}_{L}$, Design Description: $\mathbb{T}$.}
    \State {$\mathcal{G}_{\mathcal{D}^{Q}} \gets \Omega(\mathcal{D}^{Q})$} \Comment{graph construction, Sec \ref{sec:grad}}
    \For{$\mathcal{D}^{\text{IC}}_{k} \in \mathcal{D}^{\text{IC}}$}
        \State {$\mathcal{G}_{\mathcal{D}^{\text{IC}}_{k}} \gets \Omega(\mathcal{D}^{\text{IC}}_{k})$} \Comment{graph construction, Sec \ref{sec:grad}}
        \State {$\mathbb{S}_{\text{l}} \gets \alpha \mathbb{D}_{\text{W}}(\mathcal{G}_{\mathcal{D}^{Q}},\mathcal{G}_{\mathcal{D}^{\text{IC}}_{k}}) + (1 - \alpha) \mathbb{D}_\mathrm{GW}(\mathcal{G}_{\mathcal{D}^{Q}},\mathcal{G}_{\mathcal{D}^{\text{IC}}_{k}})$} \Comment{Sec \ref{sec:grad}}
        \State {$\mathbb{S}_{\text{g}} \gets 1 - \texttt{cos}(\text{CLIP}(\mathcal{D}^{Q}), \text{CLIP}(\mathcal{D}^{\text{IC}}_{k}))$}  \Comment{Sec \ref{sec:grad}}
        \State {$\mathbb{S}(\mathcal{D}^{Q},\mathcal{D}^{\text{IC}}_k) \gets \mathbb{S}_{\text{l}} + \mathbb{S}_{\text{g}}$}  \Comment{Sec \ref{sec:grad}}
        \State {$\mathbb{S}_{L} \gets \texttt{\textbf{Insert}}(\mathbb{S}_{L}, \mathbb{S}(\mathcal{D}^{Q},\mathcal{D}^{\text{IC}}_k))$} \Comment{inserting scores to list}
    \EndFor
    \State {$\mathbb{I}_{S} \gets \texttt{\textbf{Argsort}}(\mathbb{S}_{L})$} \Comment{index sorting}
    \State {$\mathbb{I}_{K} \gets \mathbb{I}_{S}[:K]$} \Comment{Select top-\textit{K} indices}
    \State {$\mathbb{T}_{\mathcal{D}^{Q}} \gets \Psi(\mathcal{D}^{Q}, \mathcal{BB}^{Q})$} \Comment{description generation, Sec \ref{sec:sdd}}
    \State {$A_D, B_{A_D}, N_D \gets A_M(\mathcal{D}^{Q},\mathbb{T}_{\mathcal{D}^{Q}},A_S,B_{A_S},N_S,\mathcal{W})$} \Comment{planning}
    \For {$n_s \in {1,\cdots,N_S}$} \Comment{reviewing, Sec \ref{sec:adrs}}
        \State {$\mathcal{R}_{S}(n_s), \mathcal{F}_{S}(n_s) \gets A_S^{(n_s)}(\mathcal{D}^{Q}, \mathbb{T}_{\mathcal{D}^{Q}}, \mathcal{D}^{\text{IC}}[\mathbb{I}_{K}], B_{A_S}^{(n_s)})$}
    \EndFor
    \For {$n_s \in {1,...,N_D}$} \Comment{reviewing, Sec \ref{sec:adrs}}
        \State {$\mathcal{R}_{D}(n_d), \mathcal{F}_{D}(n_d) \gets A_D^{(n_d)}(\mathcal{D}^{Q}, \mathbb{T}_{\mathcal{D}^{Q}}, \mathcal{D}^{\text{IC}}[\mathbb{I}_{K}], B_{A_D}^{(n_d)})$}
    \EndFor
    \State {$\mathcal{R}, \mathcal{F} \gets A_M([\mathcal{R}_{S}, \mathcal{R}_{D}], [\mathcal{F}_{S}, \mathcal{F}_{D})]$} \Comment{summarization, Sec \ref{sec:adrs}}
    \State \Return {$\mathcal{R}, \mathcal{F}$}
\end{algorithmic}
\end{algorithm}

The combined dissimilarity (since we are computing distances) score $\mathbb{S}_{\text{l}}$ is a weighted combination of Eqs. \ref{eq:wd} and \ref{eq:gwd}, based on which top-$K$ designs are selected from the in-context design library $\mathcal{D}^{\text{IC}} \subset \mathcal{D}$.
\begin{equation}
    \mathbb{S}_{\text{l}} = \alpha \mathbb{D}_{\text{W}}(\mathcal{G}_{\mathcal{D}^{Q}},\mathcal{G}_{\mathcal{D}^{\text{IC}}_{k}}) + (1 - \alpha) \mathbb{D}_\mathrm{GW}(\mathcal{G}_{\mathcal{D}^{Q}},\mathcal{G}_{\mathcal{D}^{\text{IC}}_{k}})
\end{equation}
\noindent Along with $\mathbb{S}_{\text{l}}$ (on local representations), we add global scores on renditions: $\mathbb{S}_{\text{g}} = 1 - \texttt{cos}(\text{CLIP}(\mathcal{D}^{Q}), \text{CLIP}(\mathcal{D}^{\text{IC}}_{k}))$ where \texttt{cos}($\cdot$) is the cosine similarity function. The final expression becomes: $\mathbb{S}(\mathcal{D}^{Q},\mathcal{D}^{\text{IC}}_k) = \mathbb{S}_{\text{l}} + \mathbb{S}_{\text{g}}$.

\subsection{Structured Design Description (SDD)}\label{sec:sdd}

For the input query design $\mathcal{D}^{Q}$, our goal is to generate textual descriptions $\mathbb{T}_{\mathcal{D}^{Q}}$ which contains description of elements (images, icons, texts, etc.) and how they are structured hierarchically. For example, a textual description may look like: ``A title `ABC' at the top [$bb_{11}^{Q}, bb_{12}^{Q}, bb_{13}^{Q}, bb_{14}^{Q}$] with an image of `X' below it [$bb_{21}^{Q}, bb_{22}^{Q}, bb_{23}^{Q}, bb_{24}^{Q}$]. Below the image, there is a text contaning `DEF', ... ". Passing both the graphic design and a textual description with bounding box ($\mathcal{BB}^{Q} = \{bb_{ij}^{Q}\}$) information (optional, as available from the layout metadata) improves design attribute understanding and anomaly detection by combining visual perception with explicit structural and semantic details along with element relationships. Such textual descriptions are superior to merely feeding the raw metadata information (e.g., xml or json) and assists in anchoring the MLLM responses in a detailed visual description. In addition, it makes design attribute understanding more robust to diverse layouts while facilitating clearer, more actionable feedback, reducing hallucinations. Notably, we obtain the descriptions by passing the designs through an MLLM $\Psi(\cdot)$ with suitable prompt.

\vspace{-10pt}
\begin{equation}
    \mathbb{T}_{\mathcal{D}^{Q}} = \Psi \left( \mathcal{D}^{Q}, \mathcal{BB}^{Q} \right); \text{if} \; \mathcal{BB}^{Q} \neq \emptyset, \; \text{else} \; \Psi \left( \mathcal{D}^{Q}\right)
\end{equation}

\begin{table*}[!t]
\centering

\small
\renewcommand{\arraystretch}{0.9}
\setlength{\tabcolsep}{4pt}
\resizebox{\linewidth}{!}{\begin{tabular}{@{}l |c c c|c c c|c c c|c c c @{}}
\toprule

\multirow{3}{*}{\bf Method} & \multicolumn{9}{c |}{\textit{Discrete Evaluation (Classification)}} & \multicolumn{3}{c }{\textit{Continuous Evaluation (Correlation)}}\\
\cmidrule{2-13}
& \multicolumn{3}{c |}{\bf Afixa} & \multicolumn{3}{c |}{\bf Infographic} & \multicolumn{3}{c |}{\bf \ourdataset} & \multicolumn{3}{c}{\bf GDE \cite{gde}}\\ 

\cmidrule{2-13}

& Acc $\uparrow$ & Sens $\uparrow$ & Spec $\uparrow$ & Acc $\uparrow$ & Sens $\uparrow$ & Spec $\uparrow$ & Acc $\uparrow$ & Sens $\uparrow$ & Spec $\uparrow$ & Alignment $\uparrow$ & Overlap $\uparrow$ & Whitespace $\uparrow$ \\

\midrule

Heuristic-based Evaluation &  - &  - &  - &  - &  - &  - &  - &  - &  - & 0.310 & 0.476 & 0.233 \\

\midrule

Gemini-1.5-Pro & 59.45 & 62.11 & 60.18 & 54.88 & 55.61 & 54.85 & 64.37 & 67.89 & 65.59 & 0.586 & 0.759 & 0.641 \\

Gemini-1.5-Pro + GRAD & 62.19 & 64.08 & 62.45 & 56.21 & 60.72 & 56.18 & 68.21 & 68.65 & 67.16 & 0.623 & 0.778 & 0.676 \\

Gemini-1.5-Pro + GRAD + SDD & 65.62 & 67.95 & 66.31 & 59.76 & 63.84 & 59.05 & 69.09 & 68.94 & 70.42 & 0.671 & 0.783 & 0.691 \\

\rowcolor{Light}

\bf \ourapproach$_{\text{Gemini-1.5-Pro}}$ & 72.17 & 74.94 & 70.85 & 65.97 & 67.41 & 68.58 & 75.43 & \bf 75.31 & 76.22 & 0.712 & 0.821 & 0.739 \\

\midrule

\textcolor{blue}{$\Delta_{\text{\ourapproach - Gemini-1.5-Pro}}$} & \textcolor{blue}{12.72} \textcolor{blue}{$\uparrow$} & \textcolor{blue}{12.83} \textcolor{blue}{$\uparrow$} & \textcolor{blue}{10.67} \textcolor{blue}{$\uparrow$} & \textcolor{blue}{11.09} \textcolor{blue}{$\uparrow$} & \textcolor{blue}{11.80} \textcolor{blue}{$\uparrow$} & \textcolor{blue}{13.73} \textcolor{blue}{$\uparrow$} & \textcolor{blue}{11.06} \textcolor{blue}{$\uparrow$} & \textcolor{blue}{7.42} \textcolor{blue}{$\uparrow$} & \textcolor{blue}{10.63} \textcolor{blue}{$\uparrow$} & \textcolor{blue}{0.126} \textcolor{blue}{$\uparrow$} & \textcolor{blue}{0.062} \textcolor{blue}{$\uparrow$} & \textcolor{blue}{0.098} \textcolor{blue}{$\uparrow$}\\

\midrule

GPT-4o  & 62.91 & 65.42 & 64.26 & 58.26 & 61.92 & 56.74 & 65.72 & 65.38 & 66.57 & 0.597 & 0.782 & 0.665 \\

GPT-4o + GRAD  & 64.57 & 68.65 & 65.18 & 60.41 & 63.57 & 59.66 & 68.51 & 67.26 & 70.85 & 0.639 & 0.796 & 0.688 \\ 

GPT-4o + GRAD + SDD & 67.33 & 69.60 & 68.21 & 64.95 & 66.21 & 62.12 & 70.16 & 69.44 & 73.92 & 0.677 & 0.809 & 0.703 \\

\rowcolor{Light}

\bf \ourapproach$_{\text{GPT-4o}}$ & \bf 75.29 & \bf 77.65 & \bf 72.53 & \bf 69.53 & \bf 75.37 & \bf 71.94 & \bf 76.78 & 74.56 & \bf 80.31 & \bf 0.722 & \bf 0.834 & \bf 0.748 \\

% \bf 84.28 & \bf 88.65 & \bf 80.53 & \bf 69.53 & \bf 75.37 & \bf 71.94 & \bf 80.78 & \bf 78.56 & \bf 84.31 & \bf 0.804 & \bf 0.852 & \bf 0.778\\

\midrule

\textcolor{blue}{$\Delta_{\text{\ourapproach - GPT-4o}}$} & \textcolor{blue}{12.38} \textcolor{blue}{$\uparrow$} & \textcolor{blue}{12.23} \textcolor{blue}{$\uparrow$} & \textcolor{blue}{8.27} \textcolor{blue}{$\uparrow$} & \textcolor{blue}{11.27} \textcolor{blue}{$\uparrow$} & \textcolor{blue}{13.45} \textcolor{blue}{$\uparrow$} & \textcolor{blue}{15.20} \textcolor{blue}{$\uparrow$} & \textcolor{blue}{12.28} \textcolor{blue}{$\uparrow$} & \textcolor{blue}{9.18} \textcolor{blue}{$\uparrow$} & \textcolor{blue}{13.74} \textcolor{blue}{$\uparrow$} & \textcolor{blue}{0.125} \textcolor{blue}{$\uparrow$} & \textcolor{blue}{0.052} \textcolor{blue}{$\uparrow$} & \textcolor{blue}{0.083} \textcolor{blue}{$\uparrow$}\\

\bottomrule
\end{tabular}}
\caption{\textbf{Performance of \ourapproach on the \ourbenchmark}. \ourapproach outperforms baseline methods by substantial margins across all metrics in both evaluation protocols (\textit{discrete} for attribute classification, and \textit{continuous} for correlation with human labels).}\label{tab:main_table}
%\vspace{-4mm}
\end{table*}

\subsection{Agentic-Design Review System}\label{sec:adrs}
% \snag{Work In Progress}
% While evaluating a design, motivated by the studies from design literature \cite{} there are significant emphasis on alignment, overlap, typography, etc. Any review system should absolutely look for these. Towards that we create a set of agents which look at each one of these identified design attributes explicitly (their use case is defined) as their duty - we defined them as static agents with fixed roles thus static agents. Further, contextualized to each designs, there might be some design specific dynamic attributes such as relative spacing, grouping, semantic affectiveness of communication, stylistic attributes, these things are contextualized on designs and are dynamic. In order to capture these, we propose a novel dynamic set of design agents. Then given a design we should identify whichever static agents and dynamic agents is spawned, this decision is taken care by the meta agent. Also, once this independent agents give the rating, there should be someone to aggregate these ratings, come to a consensus and generate a unified feedback by consolidating the reviews and scores for each reviewer by considering the actionable constructive feedback which can potentially improve these designs. We refer the former process as planning and latter process summarization.

When evaluating a design, insights from design literature \cite{graham2002basics, carpenter2019design, williams2007non} highlight the critical importance of attributes such as alignment, overlap and spacing.
% \balaji{{if we expect the agents to be experts in aspects, we need to have a clear enumeration of these. Words like `\etc' indicate some level of indecision here. }}
Any effective review system must rigorously assess these foundational principles to ensure high-quality design evaluation. To achieve this, we introduce a structured, agentic review framework \ourapproach where specialized agents focus on specific design attributes. Consequently, we introduce a set of \textit{Static Agents} (evaluators/reviewers), which are predefined and operate with fixed (static) roles, each dedicated to assessing well-established design attributes that are universally relevant across designs — such as alignment, overlap and spacing. These agents serve as the backbone of the evaluation system, ensuring that essential design attributes are consistently reviewed.

However, design evaluation is not solely a rule-based exercise; context-dependent attributes often influence how a design is perceived and interpreted. To capture these contextual nuances, we introduce a novel set of \textit{Dynamic Agents} (evaluators/reviewers), which adapt based on the unique characteristics of a given design. These agents assess factors such as relative spacing, grouping, semantic effectiveness of communication, stylistic coherence, etc. which vary across different designs (i.e., dynamic in nature and contextualized on designs) and are not universally predefined. 

The informed decision of which static and dynamic agents should be activated for a particular query design is exclusively handled by a \textit{Meta Agent} (evaluators/reviewers), which intelligently plans the evaluation process, ensuring that the most relevant aspects of a design are meticulously scrutinized - this constitutes the \textbf{\textit{planning}} phase. Once individual agents have assessed the design and provided their ratings (\textbf{\textit{reviewing}} phase), the system must aggregate their insights into a coherent and actionable review. A consolidation mechanism, also managed by the Meta Agent, performs \textbf{\textit{summarization}}, synthesizing feedback, resolving potential inconsistencies, and generating a unified evaluation report. This ensures that the final assessment is not just a collection of disjointed scores but a holistic, constructive critique that can guide meaningful improvements in the design. Notably, as opposed to prior design evaluation systems, either relying on predefined heuristics \cite{odonovan} or end-to-end learned scoring functions \cite{goyal2024design}, both of which lack adaptability, our framework introduces an adaptive mechanism, (akin to a peer-review process), making our approach suitable for evolving design principles and diverse stylistic elements.

We formally define the agents participating in the design evaluation process, comprising of Meta Agent $A_{M}$, Static Agents $A_{S_i} \forall i \in \{1,2,...,N_S\}$, and Dynamic Agents $A_{D_i} \forall i \in \{1,2,...,N_D\}$. $\mathcal{S}$ represents the state space describing the design under evaluation. The joint action space is denoted by $\mathbb{A}$, where each agent $A_i$ executes actions $a_i$ based on its policy $\pi_i$. The transition function $\mathcal{T}: \mathcal{S} \times \mathbb{A} \to \mathcal{S} $ models the evolution of the design evaluation state as agents contribute their assessments. For a given query design $\mathcal{D}^{Q}$, the agents collaborate in a structured sequence, via a directed interaction network (Fig \ref{fig:mainfig}) which can be decomposed into the aforementioned 3 phases (planning, reviewing, summarization), which we describe below.

\noindent \textbf{(i) Planning:} The meta-agent $A_M$ plans the evaluation process by acting as a router which initiates the process by assigning static reviewers $A_{S_i}$ based on predefined criteria and dynamic reviewers $A_{D_i}$ based on attributes sampled from the open-world design principles (attributes) $\mathcal{W}$ which are deemed contextual and relevant for the design to be evaluated. Notably, each static evaluator $A_{S_i}$ is responsible for evaluating a specific set of design attributes with predefined buckets as $B_{A_{S}}(i) \triangleq \{w_{A_{S}}^{(i)}(1), ..., w_{A_{S}}^{(i)}(k)\}, \; w_{A_{S}}^{(i)}(k) \in W_{A_S} \subset \mathcal{W}, \; i \in \{1, ..., N_S\}$, which are embedded in their respective prompts $p_{A_{S}}^{(i)}\left(B_{A_{S}}(i)\right) \in \mathcal{P}$ (the collective prompt space is represented as $\mathcal{P}$). Whereas, for each dynamic evaluator, attributes are first sampled $W_{A_D} \subseteq \mathcal{W} - W_{A_S}$ and dynamically bucketed on the fly (as decided by the meta agent) into $N_D$ buckets where each $B_{A_{D}}(i) \triangleq \{w_{A_{D}}^{(i)}(1), ..., w_{A_{S}}^{(i)}(l)\}, \; w_{A_{D}}^{(i)}(l) \in W_{A_D}, \; i \in \{1, ..., N_D\}$ followed by dynamic prompt creation $p_{A_{D}}^{(i)} \in \mathcal{P}$.

\begin{equation}
    a_M \sim \pi_M(\cdot | s, \mathcal{D}^{Q}, \mathcal{D}^{\text{IC}}_{K}), \quad \; s \sim \mathcal{S}
\end{equation}

\noindent \textbf{(ii) Reviewing:} Both groups of agents (static and dynamic) follow a policy $\pi_i$, determining its output based on the previous agent (i.e., meta agent), and the top-$K$ n-context (IC) samples $\mathcal{D}^{\text{IC}}_{K}$ from the in-context Data Library $\mathcal{D}^{\text{IC}}$ ($\mathcal{D}^{\text{IC}}_{K} \subseteq \mathcal{D}^{\text{IC}} \subset \mathcal{D}$), in the interaction network:
\begin{equation}
    a_i \sim \pi_i(\cdot | s, \{a_M\}_{A_M}, \mathcal{D}^{Q}, \mathcal{D}^{\text{IC}}_{K}), \quad s \sim \mathcal{S}
\end{equation}
where $A_M$'s outputs influence $A_i$. The aggregated actions $a = (a_1, \dots, a_{N})$ collectively determine the next state: $s_{t+1} = \mathcal{T}(s_t, a_t) = \text{Concat}(s_t, a_t)$,
where \( \text{Concat} \) represents the concatenation operation that updates the design evaluation context. These agents assign \textit{quantitative} measures (design attribute ratings) and \textit{qualitative} measures (actionable feedback), respectively denoted by $\mathcal{R}_S$ and $\mathcal{F}_S$ (for static agents), and $\mathcal{R}_D$ and $\mathcal{F}_D$ (for dynamic agents).

\noindent\textbf{(iii) Summarization:} The meta-agent $A_M$ collates the scores and integrates the feedback it receives from all static and dynamic reviewers to construct a final evaluation.
\begin{equation}
    a_M \sim \pi_M(\cdot | s, \{a_j\}_{A_j \in \text{Pred}(A_M)}), \quad s \sim \mathcal{S}
\end{equation}
where $\text{Pred}(A(i))$ represents predecessor agent(s) who directly influence the outputs of the successor agent(s). Using this mechanism, redundant feedback is removed, refining the assessment and obtaining a final list of actionable feedback $\mathcal{F}$ (from $\mathcal{F}_S$, $\mathcal{F}_D$) and Attribute Ratings $\mathcal{R}$ (from $\mathcal{R}_S$, $\mathcal{R}_D$).

\subsection{Overall Framework}

We summarize the overall flow of our pipeline in Algo \ref{alg:algorithm_ADRS}. Our key novelties are: (\textit{i}) to adapt the graph matching algorithm based on localized representations (Supp Algos 1 - 2) for selecting the top-$K$ designs based on scores (lines 1 - 9, Algorithm \ref{alg:algorithm_ADRS}), (\textit{ii}) to anchor MLLM responses via structured design descriptions generated using design renditions and bounding boxes (line 10, Algo \ref{alg:algorithm_ADRS}), and (\textit{iii}) to introduce agentic design review framework comprising of meta, static and dynamic agents and involving planning, reviewing and summarization mechanisms (lines 11 - 16, Algo \ref{alg:algorithm_ADRS}).

\begin{table}[!t]
\centering
\small
\renewcommand{\arraystretch}{1.0}
\setlength{\tabcolsep}{4pt}
\resizebox{\columnwidth}{!}{\begin{tabular}{@{}l | c | c c c c @{}}
\toprule
\multirow{1}{*}{\bf Method} & \multicolumn{1}{c |}{\bf Metric} & \multicolumn{1}{c }{\bf GDE} & \multicolumn{1}{c }{\bf Infographic} & \multicolumn{1}{c }{\bf Afixa} & \multicolumn{1}{c }{\bf \ourdataset} \\
\midrule
\ourapproach$_{\text{Gemini}}$ & Human & 0.744 & 0.682 & 0.720 & 0.742 \\
\midrule
\rowcolor{Light!50}
\ourapproach$_{\text{Gemini}}$ & AIM$_{\text{Sim}}$ & 0.851 & 0.806 & 0.821 & 0.834 \\
\rowcolor{Light!50}
\ourapproach$_{\text{Gemini}}$ & AIM$_{\text{Gemini}}$ & 0.792 & 0.758 & 0.769 & 0.802\\
\rowcolor{Light!50}
\ourapproach$_{\text{Gemini}}$ & AIM$_{\text{GPT-4o}}$ & 0.795 & 0.736 & 0.762 & 0.785 \\
\midrule
\ourapproach$_{\text{GPT-4o}}$ & Human & 0.762 & 0.708 & 0.736 & 0.740 \\
\midrule
\rowcolor{Light!50}
\ourapproach$_{\text{GPT-4o}}$ & AIM$_{\text{Sim}}$ & 0.881 & 0.835 & 0.817 & 0.837 \\
\rowcolor{Light!50}
\ourapproach$_{\text{GPT-4o}}$ & AIM$_{\text{Gemini}}$ & 0.829 & 0.774 & 0.794 & 0.791 \\
\rowcolor{Light!50}
\ourapproach$_{\text{GPT-4o}}$ & AIM$_{\text{GPT-4o}}$ & 0.832 & 0.763 & 0.803 & 0.804\\ 

\bottomrule
\end{tabular}}
\caption{\textbf{Feedback evaluation of \ourapproach} on both GPT-4o and Gemini versions. Strong correspondences can be observed across different variations of AIM and also with the human ratings.}
\label{tab:feedback}
% \vspace{-2mm}
\end{table}

\section{Experiments and Results}

\subsection{\ourbenchmark: Design Evaluation Benchmark}\label{sec:benchmark}

Graphic design principles provide essential guidelines for creating clear, coherent, and usable compositions \cite{carpenter2019design, graham2002basics, williams2007non}. Existing heuristic-based evaluation methods \cite{odonovan} often assess only a few principles, such as alignment, overlap, and white space, or provide a single \textit{goodness-of-fit} score \cite{goyal2024design}, overlooking key aspects like visual hierarchy, color pairings, and font compatibility. Despite advancements in automated design generation \cite{luo2024layoutllm, inoue2023layoutdm, yamaguchi2021canvasvae, yang2024posterllava, tang2023layoutnuwa, lin2023layoutprompter, guerreiro2024layoutflow}, there is no standardized evaluation framework or benchmark to capture critical design attributes. This gap hinders objective assessment, limits comparisons, and makes scaling reliable, data-driven design models challenging. Thus, we propose \ourbenchmark, a unified benchmark for evaluating design effectiveness, enabling fair comparisons, improving automated design tools, and enhancing overall design quality.

\noindent {\textbf{Attributes.}} A core aspect of our benchmark is the multi-dimensional evaluation of designs through a comprehensive set of design attributes, collectively termed “World Attributes” ($\mathcal{W}$). These attributes capture fundamental design flaws and best practices. \textit{Text-rendering quality} assesses legibility based on font size, weight, and clarity. \textit{Too many words} penalizes cluttered text, while \textit{too many fonts} and \textit{bad typography colors} flag inconsistent styling and poor contrast. \textit{Composition and layout}, \textit{alignment}, and \textit{spacing} ensure visual balance, favoring structured designs with clear margins and uniform spacing. \textit{Color harmony} and \textit{wrong color palettes} measure the effectiveness of color choices in maintaining aesthetic appeal. \textit{Style} evaluates stylistic aspects and consistency, while \textit{grouping} and \textit{image-text alignment} address content organization and readability. \textit{Aesthetics} considers overall visual appeal, penalizing \textit{overlap}, \textit{bad images}, and poorly arranged elements.

\noindent {\textbf{Datasets.}} In \ourbenchmark, we include 4 datasets which are:

\noindent \underline{\textit{GDE:}} \ourbenchmark leverages publicly available GDE dataset \cite{gde}, hosting a large collection of 700 banner and poster designs each of which containing 3 attributes: alignment, overlap, and white space, on a 1 to 10 scale. Layout metadata information are not publicly available for GDE.

\noindent \underline{\textit{Afixa:}} It consists of 71 designs, in \ourbenchmark as collected from a public platform Roboflow\cite{roboflow} which consists of \textit{yes} or \textit{no} values for 5 design attributes: \textit{wrong color palettes pairings}, \textit{bad typo colors}, \textit{bad images}, \textit{too many words}, \textit{too many fonts}. Layout metadata information are not available.

\noindent \underline{\textit{Infographic:}} It consists of 55 samples collected from the Roboflow platform. This consists of the layout metadata information (xml) for the elements (i.e., bounding boxes for the elements). Each design has \textit{yes} or \textit{no} responses for all the 15 attributes (see Sec \ref{sec:benchmark}).
% \balaji{If you are referring to design attributes that was collected for the stylize work, note that this was a result of an internal data collection. I don't think we should share this externally (including stats and details) since there is a business connection here.}

\noindent \underline{\textit{Internal Design Dataset (\ourdataset):}} We internally collect 137 design samples each of which has \textit{yes} or \textit{no} responses to the 15 design attributes (see Sec \ref{sec:benchmark}). The designs are professionally curated and typically comprises of flyers, invitations, posters, albeit with some inconsistencies in design attributes. \ourdataset consists of the layout metadata information (xml) with bounding boxes for the design elements.

\subsection{Evaluation Metrics}\label{sec:metrics}

\noindent \textbf{Attribute Evaluation:} Design flaws often coexist, impacting readability, style, and aesthetics. Assigning multiple labels allows for a more detailed evaluation rather than reducing it to a single \textit{goodness-of-fit} score. For Afixa, Infographic, and \ourdataset, attributes are labeled as \textit{yes} or \textit{no}, making it a multi-label classification problem. We report the mean Sensitivity, Specificity, and Accuracy, termed \textit{Discrete Evaluation}. For GDE, attributes are rated on a continuous scale (1–10), so we follow \cite{gde} and report Pearson correlation with human labels, termed \textit{Continuous Evaluation}.
% A single design flaw does not exist in isolation; multiple issues can coexist, affecting overall readability, style and aesthetics. Therefore, assigning multiple labels enables a richer and more detailed evaluation of design quality leading to pinpointing multiple problem areas in one evaluation, as opposed to reducing it to a single \textit{goodness-of-fit} value. For the Afixa, Infographic, and \ourdataset datasets, we have \textit{yes} or \textit{no} value across each attribute. Therefore, it reduces to multi-label classification problem and we report the mean of Sensitivity, Specificity and Accuracy across all the attributes - we term this as \textit{Discrete Evaluation}. For the GDE dataset, across attributes we have a continuous scale rating (between 1 and 10). Hence, following \cite{gde}, we report the Pearson correlation values with the human labels - we call this as \textit{Continuous Evaluation}.

\noindent \textbf{Feedback Evaluation:} To evaluate feedback quality, we use the Actionable Insights Metric (AIM) with two approaches: (\textit{i}) AIM${\text{GPT-4o}}$/AIM${\text{Gem}}$, where GPT-4o/Gemini rates how well the feedback addresses ground truth problems (converted into sentences), and (\textit{ii}) AIM$_{\text{Sim}}$, which measures semantic similarity between problems and feedback using the GTE-L model \cite{gtelarge}.
% To assess the quality of generated feedback, we employ a Actionable Insights Metric (AIM) with two approaches - (\textit{i}) AIM$_{\text{GPT-4o}}$/AIM$_{\text{Gem}}$: using a GPT-4o / Gemini-based evaluation we give the textual actionable \textit{feedback} along with the ground truth data (\textit{problems}) converted to sentences (e.g., \{`too many colors': yes, `related elements not grouped': yes\} $\rightarrow$ `The problem(s) in design is/are: the design has too many colors and related elements are not grouped.'), and ask GPT-4o/Gemini `Is the feedback a reasonable solution for the problems? Give a rating out of 10.', (\textit{ii}) AIM$_{\text{Sim}}$: using a GTE-L model \cite{gtelarge} for computing semantic similarities between the problem(s) and the actionable feedback/solution texts.

\subsection{Baselines}

To the best of our knowledge, there does not exist any design-aware MLLMs let alone agentic frameworks. Hence, to compare \ourapproach, we introduce some baselines with two powerful MLLMs, Gemini-1.5 Pro \cite{geminipro} and GPT-4o \cite{gpt4o}, and subsequently adding design-aware components such as GRAD and SDD. However, these baselines are exclusively restricted to a single MLLM agent. Furthermore, notably, vanilla GPT-4o implementation for design has been proposed in \cite{gde} which we adopt in Table \ref{tab:main_table}. Along with these, we also report heuristic evaluation results following \cite{odonovan} on the GDE dataset \cite{gde} only. Afixa, Infographic and \ourdataset have design attributes which go well beyond conventional heuristics (e.g., alignment, overlap and spacing) and therefore cannot be evaluated with the heuristics-based method.

\begin{figure}[!t]
    \centering
    \includegraphics[width=\linewidth]{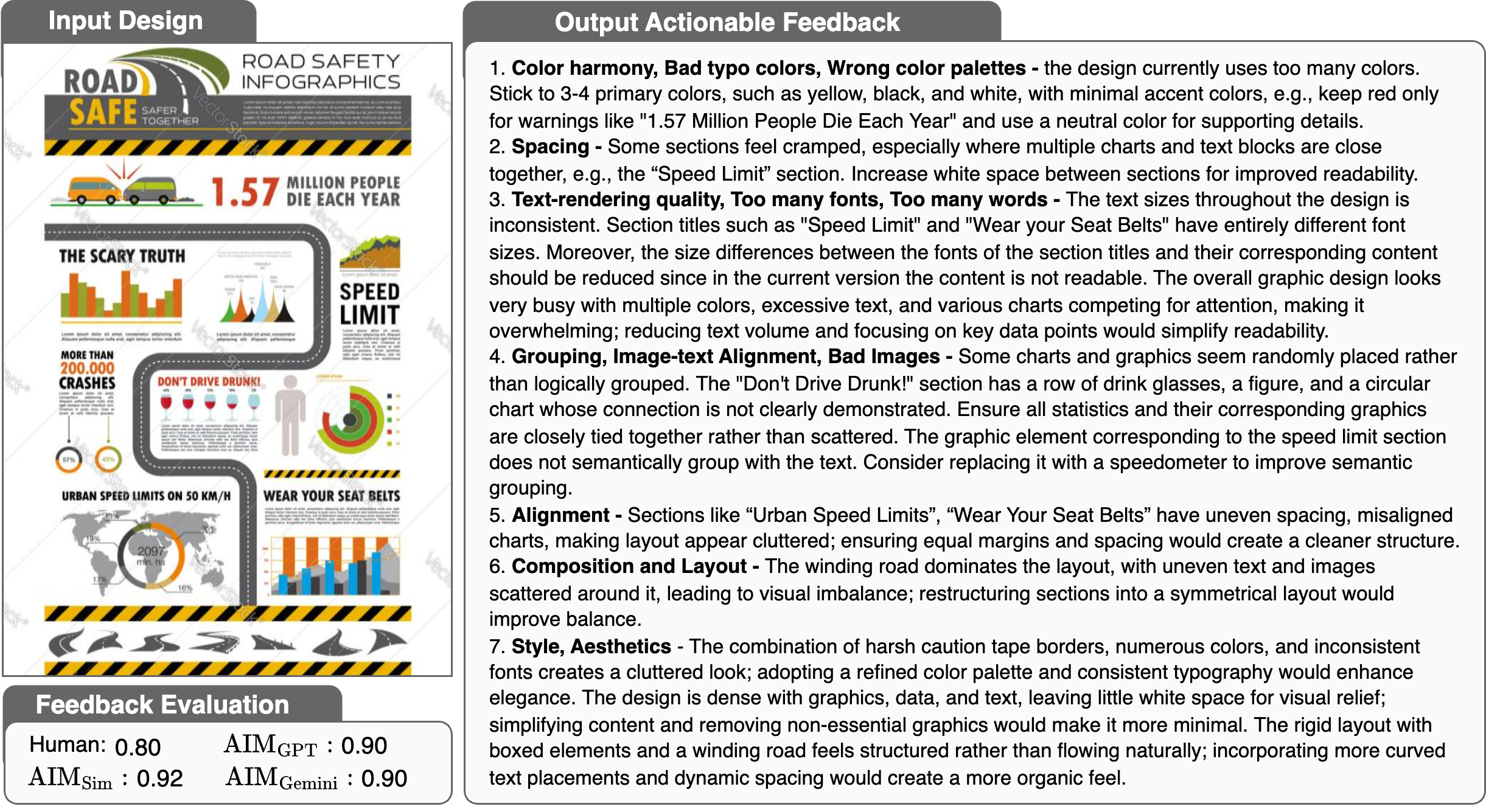}
    \caption{\textbf{Generated feedback along with the design attributes} which are found to be inconsistent for the input design evaluated (best viewed when zoomed). Feedback evaluation scores are also reported (as outlined in Sec \ref{sec:metrics}).}
    \label{fig:qual_fig_1}
\end{figure}

\begin{table}[!t]
\centering
\small
\renewcommand{\arraystretch}{0.8}
\setlength{\tabcolsep}{4pt}
\resizebox{\columnwidth}{!}{\begin{tabular}{@{}l | c c c @{}}
\toprule
\multirow{1}{*}{\bf In-context Sampling Method} & \multicolumn{1}{c }{\bf GDE (Avg.)} & \multicolumn{1}{c }{\bf Infographic (Acc)} & \multicolumn{1}{c }{\bf Afixa (Acc)} \\
\midrule

Random Selection (9) & 0.743 & 63.75 & 72.06\\
Global Features (6) & 0.752 & 64.39 & 72.94\\
Description-based (6) & 0.755 & 67.05 & 73.98\\
\midrule
\rowcolor{Light}
GRAD (w/o Global Features) (5) & 0.760 & 67.13 & 74.81\\
\rowcolor{Light}
\bf GRAD (4) & \bf 0.768 & \bf 69.53 & \bf 75.29\\

\bottomrule
\end{tabular}}
\caption{\textbf{Ablation on different in-context design selection methods} for \ourapproach$_{\text{GPT-4o}}$. Best results are obtained with GRAD, preserving semantic, spatial and structural information. Best $K$ for top-$K$ retrieval are provided in brackets beside method.}\label{tab:ablation_in_context}
% \vspace{-2mm}
\vspace{-10pt}
\end{table}

\begin{table}[!t]
\centering
\small
\renewcommand{\arraystretch}{0.6}
\setlength{\tabcolsep}{4pt}
\resizebox{\columnwidth}{!}{\begin{tabular}{@{}c | c c | c | c @{}}
\toprule
\multirow{2}{*}{\bf GRAD} & \multicolumn{2}{c |}{\bf SDD} & \multirow{2}{*}{\bf Infographic (Acc)} & \multirow{2}{*}{\bf \ourdataset (Acc)} \\
\cmidrule{2-3}
& w/ Bnd. Box & w/o Bnd. Box & & \\
\midrule

%\cmidrule{2-16}

%F-$1$ & F-$1$ & F-$1$ & F-$1$ & F-$1$ \\

\textcolor{OrangeRed}{\ding{55}} & \textcolor{OrangeRed}{\ding{55}} & \textcolor{OrangeRed}{\ding{55}} & 63.11 & 69.75 \\

\textcolor{OrangeRed}{\ding{55}} & \textcolor{ForestGreen}{\ding{51}} & \textcolor{OrangeRed}{\ding{55}} & 64.52 & 71.03 \\ 

\textcolor{OrangeRed}{\ding{55}} & \textcolor{OrangeRed}{\ding{55}} & \textcolor{ForestGreen}{\ding{51}}  & 65.71 & 72.86 \\

\textcolor{ForestGreen}{\ding{51}} & \textcolor{OrangeRed}{\ding{55}} & \textcolor{OrangeRed}{\ding{55}} & 66.48 & 73.95 \\

\textcolor{ForestGreen}{\ding{51}} & \textcolor{ForestGreen}{\ding{51}} & \textcolor{OrangeRed}{\ding{55}} & 67.82 & 74.57 \\

\rowcolor{Light}
\textcolor{ForestGreen}{\ding{51}} & \textcolor{ForestGreen}{\ding{51}} & \textcolor{ForestGreen}{\ding{51}} & \bf 69.53 & \bf 76.78 \\

\bottomrule
\end{tabular}}
\vspace{-10pt}
\caption{\textbf{Effect of different components} of \ourapproach$_{\text{GPT-4o}}$. Results on dataset with layout metadata information are reported. Substantial improvements are found upon adding GRAD and SDD.}\label{tab:ablation_components}
\end{table}

\begin{table}[!t]
\begin{minipage}[b]{0.48\linewidth}
    \centering
    % % \vspace{-15mm}
    \renewcommand{\arraystretch}{0.75}
    \setlength{\tabcolsep}{4pt}
    \resizebox{\linewidth}{!}{\begin{tabular}{@{}c | c c@{}}
    \toprule
    \multirow{1}{*}{\bf $\alpha$} & \multicolumn{1}{c }{\bf Infog. (Acc)} & \multicolumn{1}{c }{\bf Afixa (Acc)} \\
    \midrule
    
    0  & 68.10 & 73.95 \\
    1 & 66.85 & 72.86 \\
    0.25 & 68.94 & 74.38 \\
    0.75 & 67.21 & 73.14 \\
    \rowcolor{Light}
    \bf 0.5 & \bf 69.53 & \bf 75.29 \\
    
    \bottomrule
    \end{tabular}}
        % \vspace{-3.0mm}
        \captionof{table}{\textbf{Ablation on $\alpha$} for \ourapproach$_{\text{GPT-4o}}$ on Infographic and Afixa datasets.}
    \label{tab:ablation_alpha}
  \end{minipage}
  \hfill
  \begin{minipage}[b]{0.49\linewidth}
    \centering
    % \includegraphics[width=0.9\linewidth]{ECCV/Figures/Ablations_plot.png}
    % % \vspace{-5mm}
	\centering
    %\hspace{-1.0em}
        \begin{subfigure}[b]{\linewidth}
		\resizebox{!}{!}{
			\begin{tikzpicture}
	\begin{axis} [
         width=\textwidth,
         height=.8\textwidth,
		axis x line*=bottom,
		axis y line*=left,
		legend pos=north east,
		ymin=55, ymax=71,
		xmin=0, xmax=6,
		xticklabel={\pgfmathparse{\tick}\pgfmathprintnumber{\pgfmathresult}},
		xtick={1,2,4,6},
		ytick={64,66,68,70},
        xlabel={\footnotesize{$K$}},
        ylabel={\footnotesize{Accuracy}},
        xlabel shift=-5pt,
        ylabel shift=-5pt,
		width=\linewidth,
		legend style={cells={align=left}},
		label style={font=\scriptsize},
		tick label style={font=\scriptsize},
		legend style={draw=none,at={(0.30,0.35)},anchor=west},
		]

        \addplot[mark=*,mark options={scale=0.5, fill=Dark},style={thick},Dark] plot coordinates {
			(1, 67.1)
                (2, 67.9)
			(4, 69.5)
			(6, 69.4)
		};
        \addlegendentry{\scriptsize{\textsc{A-DRS}$_{\text{GPT}}$}}

        \addplot[mark=*,mark options={scale=0.5, fill=gray},style={thick},gray] plot coordinates {
			(1, 63.5)
                (2, 65.1)
			(4, 66.0)
			(6, 66.0)
		};
        \addlegendentry{\scriptsize{\textsc{A-DRS}$_{\text{Gem}}$}}

		% \addplot[dashed,gray] plot coordinates {
		% 	(0, 26.0)
		% 	(100, 26.0)
		% };
		% \addplot[mark=asterisk,color_ao,mark options={scale=2,thick},only marks] plot coordinates {
		% 	(1, 67.26)
		% };
		% \label{pgf:ek100_map}
	\end{axis}
\end{tikzpicture}%
		}
        \label{fig:ablationK}
        \end{subfigure}
    \captionof{figure}{\textbf{Impact of $K$ in GRAD} for \ourapproach.}
  \end{minipage}
  \vspace{-10pt}
\end{table}

\subsection{Results}

Table \ref{tab:main_table} presents quantitative comparisons of \ourapproach against baseline methods on \ourbenchmark. Each MLLM-based evaluation reports the mean of five independent runs. Our results highlight the benefits of incorporating design awareness and multi-agent interaction, significantly outperforming single-agent systems in both Discrete (multi-attribute classification) and Continuous (correlation with human labels) evaluations. Notably, GPT-4o consistently outperforms Gemini-1.5-Pro across all datasets and metrics (except Sensitivity on \ourdataset), aligning with prior findings \cite{zhou2024image, qi2023gemini}. On the GDE dataset, MLLM-based evaluations (both single-agent and \ourapproach) achieve significant gains over heuristic-based methods, corroborating \cite{gde}. Table \ref{tab:feedback} reports qualitative scores for \ourapproach${\text{Gemini}}$ and \ourapproach${\text{GPT-4o}}$ on four datasets, using the three feedback evaluation metrics from Sec \ref{sec:metrics}. For reference, we include Human scores, averaged from five raters and normalized between 0 and 1. AIM$_{\text{Sim}}$, AIM$_{\text{Gemini}}$, and AIM$_{\text{GPT-4o}}$ scores strongly correlate with Human ratings, validating the effectiveness of AIM for actionable feedback evaluation.

Fig. \ref{fig:qual_fig_1} presents a qualitative example from the Infographics dataset, showcasing how \ourapproach generates actionable feedback for design evaluation. The feedback quality is validated by both AIM scores (Sec \ref{sec:metrics}) and human ratings, normalized between 0 and 1 for fair comparison. Our GRAD module selects relevant samples for design understanding, while SDD grounds element relative positions, enhancing evaluation insights. This enables references to specific design elements (e.g., inconsistent spacing in the "Speed Limit" section), improving feedback \textit{actionability}. Additionally, stylistic aspects like elegance and minimality are correctly assessed by dynamic experts, spawned appropriately by the meta agent.

\section{Discussions and Analysis}

\subsection{Analyzing the Impact of GRAD}

Table \ref{tab:ablation_in_context} compares GRAD with four retrieval approaches. A \textit{random} selection of exemplar designs yields the lowest scores on \ourbenchmark. Using CLIP \textit{global features} improves performance over random selection, consistent with prior findings \cite{ferber2024context}. A \textit{description-based} approach, which retrieves top-$K$ samples using SDD-generated textual descriptions (Sec \ref{sec:sdd}), outperforms global features, especially when design element descriptions with coordinates are available (e.g., Infographic). Removing global features from GRAD (Sec \ref{sec:grad}) leads to a slight performance drop, highlighting the need for both structural and global features.

GRAD achieves the best performance while requiring fewer in-context samples, as it preserves spatial and compositional relationships rather than treating designs as unordered features. Figure \textcolor{red}{4} shows similar trends in Accuracy across different $K$ values for GPT-4o and Gemini on the Infographic dataset, demonstrating GRAD's robustness. Finally, Table \ref{tab:ablation_components} shows that even without SDD, GRAD significantly boosts performance, as seen in Table \ref{tab:main_table} for single MLLMs.

\subsection{Assessing the Role of SDD}
Design renditions capture aesthetic and spatial patterns, while textual descriptions provide an anchor for understanding. Table \ref{tab:ablation_components} shows that even without GRAD, SDD significantly boosts evaluation performance in \ourapproach, with similar trends in Table \ref{tab:main_table} for single MLLM models (GPT/Gemini). Performance improves further with bounding boxes, demonstrating that detailed visual descriptions with coordinate references (akin to phrase grounding or referring expressions) enhance model comprehension and anomaly detection in designs. We use structured descriptions over raw XML metadata because LLMs inherently process textual information more effectively. % (see Supplementary).
% While design renditions allow models to capture aesthetic and spatial patterns, textual descriptions help to simplify their understanding by providing an anchor. Consequently, in Table \ref{tab:ablation_components}, where we observe that even in the absence of GRAD, SDD contributes to substantial boost in evaluation performance in \ourapproach (likewise observations can be noted in Table \ref{tab:main_table} for a single MLLM (GPT/Gemini) model). However, notably, in the presence of bounding boxes of design elements, the performance further improves, demonstrating that detailed visual descriptions along with their coordinate references (akin to phrase grounding or referring expressions) can enhance the comprehension capabilities of these models enabling the detection of anomalies in the designs. Finally, the reason we chose a structured description over passing the raw metadata information as jsons or xmls is because the backbone of these models are LLMs and they fathom textual descriptions better than other forms of expression (as also shown in Supplementary).

\subsection{Analyzing the Effect of $\alpha$} 
Table \ref{tab:ablation_alpha} shows the impact of different $\alpha$ values in balancing Wasserstein (WD) and Gromov-Wasserstein distances (GWD). Using only GWD ($\alpha = 0$) outperforms WD alone ($\alpha = 1$), but the best performance occurs at $\alpha = 0.5$, denoting the importance of combining semantic feature matching with structural similarities for improved in-context retrieval.

% In Table \ref{tab:ablation_alpha}, we report the effect of choosing different $\alpha$ values which balances the importance of Wasserstein (W) distance and Gromov-Wasserstein (GW) distance. Using only GW-distance ($\alpha = 0$) is better as compared to using only W-distance ($\alpha = 1$), as seen from the results in Table \ref{tab:ablation_alpha}. However, leveraging both with $\alpha = 0.5$ yields the best performance out of the different values reported, highlighting the importance of semantic feature matching alongside structural similarities for better in-context retrieval.

% \begin{tcolorbox}[colback=Light!50, colframe=gray!80]
% \includegraphics[width=0.3cm]{Figures/lightbulb_logo.png} {\small \ourapproach ensures structured, context-aware evaluation of designs alongside generating actionable feedback and can adapt across diverse design styles.}

% \includegraphics[width=0.3cm]{Figures/lightbulb_logo.png} {\small Design awareness can be injected via GRAD and SDD modules which improve overall evaluation performance.}  
% \end{tcolorbox}

\section{Conclusion}

We propose the first Agentic framework for critiquing a graphic design across multiple dimensions to generate scores and actionable feedback. This can serve as a tool in empowering novice designers to enhance their designs, and for evaluating the quality of designs generated by generative models \cite{jia2023cole,inoue2024opencole} alike. Our novel exemplar selection approach and prompt expansion technique is critical towards developing the framework, as validated through our extensive experimental analysis. A key next step would be to extend the framework to automatically applying the generated actionable feedback to the input design, thereby enabling a self-improving framework for graphic design generation. 
% We trust that our work would gather more attention to this pragmatic, yet under-explored research direction.

% \joseph{Work In Progress.}

{
    \small
    \bibliographystyle{ieeenat_fullname}
    \bibliography{main}
}

\appendix

% \section{More Details on Methods}
\begin{center}
\Large\textbf{Appendix}
\end{center}

\section{More Related Works}

\paragraph{Hallucination Mitigation in MLLMs.} Hallucination is a challenge when generating content with multi-modal large models. In \citet{DBLP:journals/corr/abs-2402-00253}, the authors attribute hallucination in MLLMs to misaligned generation with the corresponding visual information. Many researchers have evaluated the density of the hallucination for MLLMs, including \citet{DBLP:conf/emnlp/LiDZWZW23} and \citet{DBLP:journals/corr/abs-2310-05338}. Commonly, hallucination mitigation is performed either by finetuning or by heuristic based post-processing \cite{DBLP:conf/aaai/GunjalYB24, DBLP:journals/corr/abs-2309-04041, DBLP:conf/iclr/LiuLLWYW24, DBLP:conf/iclr/ZhouCYZDFBY24}.

\paragraph{Multi-modal Learning.} Over the past few years, conventional multimodal approaches, spanning vision–language systems \cite{clip, unitab, uniter, glip, imagebind, volta, blip, safari, albef, egovlp, egovlpv2, maple, apollo}, audio–visual methods \cite{imagebind, avseg, chowdhury2024melfusion, avtrustbench, avsegformer, aurelia, egoadapt}, and audio–language models \cite{clap, audioclip, beats, wu2022wav2clip, merlot}, have advanced substantially, largely targeting a broad range of coarse-grained tasks (e.g., question answering, captioning, and retrieval) as well as fine-grained problems (e.g., detection, segmentation, phrase grounding, and related understanding/generation objectives). Despite this progress, such traditional architectures generally do not address reasoning-centric tasks, with NLVR being a notable exception. More recently, the emergence of multimodal large language models and agentic systems \cite{llava, instructblip, xinstructblip, pgvideollava, chowdhury2024meerkat, instructgpt, vistallm, minigpt4, minigptv2, shikra, mplugowl, macawllm, anygpt, nextgpt, anymal, avicuna, baycat, videosalmonn, bubogpt, chatbridge, cheng2024videollama, cogvlm, ying2024internlm, vita, videochatgpt, lisa, gpt4roi, kosmos, timechat, unifiedio, imagebindllm, openai_gpt4o, magnet} has prompted initial attempts to harness LLM-style reasoning for challenging multimodal question answering via multi-step inference; however, adapting these kinds of architectures to design understanding and evaluation settings - particularly in a way that produces actionable feedback remains comparatively under-explored.

\section{Algorithms for Wasserstein and GW Distances}

We provide the algorithms for Wasserstein and Gromov Wasserstein Distance computation respectively in Algorithhms \ref{algo:wass_dist} and \ref{algo:gromov_wass_dist} following \cite{cuturi2013sinkhorn, peyregromov, alvarez2018gromov, peyre2019ot}.

\begin{algorithm}[!h]
\footnotesize
\caption{Wasserstein Distance Computation in GRAD}
\label{algo:wass_dist}
\begin{algorithmic}[1]
\Require{Initial Transport Plan: $\mathbf{\Phi}^{(1)} = \mathbf{1} \mathbf{1}^\top$; Initial scaled unity matrix: $\boldsymbol{\sigma} = \frac{1}{m}\mathbf{1_m}$; Cost (similarity) matrix b/w node pairs of two graphs: $\mathbf{C}_{ij}$; Total Optimal Transport steps: $\mathcal{N}_{\mathbf{\Phi}}$;  Cost matrix decay factor: $\beta$; Scaled cost matrix: $\mathbf{\Upsilon}_{ij} = {\rm e}^{-\frac{\mathbf{C}_{ij}}{\beta}}$.}
\Ensure{Optimal Transport Plan: $\mathbf{\Phi}$; Wasserstein Distance: $\mathbb{D}_{\text{W}}$.}
\For{$t \in \{1,2,3, \cdots \mathcal{N}_{\mathbf{\Phi}}\}$}
\State $\mathbf{Q} \gets \mathbf{\Upsilon} \odot \mathbf{\Phi}^{(t)}$ \Comment{$\odot$ is Hadamard product}
    \For{$l \in \{1, \cdots, L\}$}
        \State $\boldsymbol{\delta} \gets \frac{1}{m\mathbf{Q}{\boldsymbol{\sigma}}}$, $\boldsymbol{\sigma} \gets \frac{1}{m\mathbf{Q}^\top\boldsymbol{\delta}}$
    \EndFor
    \State $\mathbf{\Phi}^{(t+1)} \gets \text{diag}(\boldsymbol{\delta})\mathbf{Q}\text{diag}(\boldsymbol{\sigma})$
\EndFor
\State $\mathbb{D}_{\text{W}} \gets \langle \mathbf{C}^{\top}, \mathbf{\Phi}\rangle$ \Comment{$\langle \cdot, \cdot \rangle$ is the Frobenius dot-product} \\
\Return $\mathbf{\Phi}$, $\mathbb{D}_{\text{W}}$
\end{algorithmic}
\end{algorithm}

\begin{algorithm}[!h]
\footnotesize
\caption{GW Distance Computation in GRAD}
\label{algo:gromov_wass_dist}
\begin{algorithmic}[1]
\Require{Nodes of Graphs $\mathcal{G}_x, \mathcal{G}_y$: $\{x_i\}_{i=1}^{m}, \{y_j\}_{j=1}^{n}$; Intra-domain Cost Matrices: $[\mathbf{C}_x]_{i,i'} = d(x_i, x_{i'}), [\mathbf{C}_y]_{j,j'} = d(y_j, y_{j'})$; probability vectors: $p = \frac{1}{m}\mathbf{1_m}, q = \frac{1}{n}\mathbf{1_n}$; Total Optimal Transport steps: $\mathcal{N}_{\hat{\mathbf{\Phi}}}$.}
\Ensure{Optimal Transport Plan: $\hat{\mathbf{\Phi}}$; GW Distance: $\mathbb{D}_{\text{GW}}$.}
% \State {$[\mathbf{C}_x]_{i,i'} \gets d(x_i, x_{i'}), [\mathbf{C}_y]_{j,j'} \gets d(y_j, y_{j'})$} \Comment{Eq. \ref{eq:edge_weight}}
\State {$\mathbf{C}_{xy} \gets \mathbf{C}_x^{2} p 1_m^{\top} + \mathbf{C}_y q (\mathbf{C}_y^2)^{\top}$}
\For{$t \in \{1,\cdots$,$\mathcal{N}_{\hat{\mathbf{\Phi}}}$\}}
    \State {$\mathcal{L} = \mathbf{C}_{xy} - 2 \mathbf{C}_x \hat{\mathbf{\Phi}}^{(t)} \mathbf{C}_y^{\top}$}
    \State {$\text{Algorithm \ref{algo:wass_dist} to compute }\hat{\mathbf{\Phi}}$}
\EndFor
\State $\mathbb{D}_{\text{GW}} \gets \langle \mathcal{L}^{\top}, \hat{\mathbf{\Phi}}\rangle$ \Comment{$\langle \cdot, \cdot \rangle$ is the Frobenius dot-product} \\
\Return $\hat{\mathbf{\Phi}}$, $\mathbb{D}_{\text{GW}}$
\end{algorithmic}
\end{algorithm}

\section{More Details on \ourbenchmark}

We group similar design attributes into the attribute buckets in Table \ref{tab:attribute_buckets} which we refer to in the main paper. These buckets are used as sets of specialized expertise of the static agents. The remaining attributes are not grouped and are present in the set of design world attributes (total 15) for the meta agent to sample from and spawn dynamic attributes contextualized on the designs to be evaluated.

\begin{tcolorbox}[float, width=\columnwidth, colback=white, colframe=Light, title=\textcolor{black}{List of Design Attributes.} ] 

Text-rendering quality, Too many words, Too many fonts, Bad typo colors, Composition and layout, Alignment, Spacing, Color harmony, Wrong color palettes, Style, Grouping, Image-text alignment, Aesthetics, Overlap, Bad images

\end{tcolorbox}

\setlength{\columnsep}{3pt}
\begin{table}[t!]
% \vspace{-20pt}
\small
\centering
\setlength{\tabcolsep}{5pt}
\renewcommand{\arraystretch}{0.25}
\resizebox{1.\columnwidth}{!}
{\begin{tabular}{c | l}
\toprule
\bf Bucket & \bf  Attributes \\
\midrule
$B_{A_S}(1)$ & Text-rendering quality, Too many words, Too many fonts, Bad typo colors \\
\midrule
$B_{A_S}(2)$ & Color harmony, Wrong color palettes, Bad typo colors.\\
\midrule
$B_{A_S}(3)$ & Composition and layout, Alignment, Spacing. \\
\bottomrule
\end{tabular}}
\caption{\textbf{Predefined Attribute Buckets} from the list of attributes. The remaining attributes are not bucketed and kept in the open world set of attributes for the meta agent to sample from and spawn dynamic attributes contextualized on the designs to be evaluated.}
\label{tab:attribute_buckets}
\end{table}

\section{More Results on \ourbenchmark}

\subsection{Results on \ourapproach$_{\text{Gemini-1.5-Pro}}$}

We provide additional results on \ourbenchmark with the Gemini-1.5-Pro \cite{geminipro} version of \ourapproach in Tables \ref{tab:ablation_in_context_gemini}, \ref{tab:ablation_components_gemini}, \ref{tab:ablation_alpha_gemini}.

\subsection{SDD vs XML metadata}

SDD approach enhances semantic understanding by providing textual descriptions of design elements and their relationships, which helps in interpreting hierarchy and structure better than raw metadata (XML). It improves robustness to layout variability by normalizing spatial relationships, and making feedback more interpretable and actionable. Additionally, it enhances multimodal reasoning by combining box information with textual descriptions, leading to better contextual decision-making. This is reflected in Table \ref{tab:ablation_sdd_xml} for both the versions of \ourapproach, where we report SDD vs. passing layout metadata in the xml format directly alongside input design, and we find SDD to be superior. However, SDD may introduce processing overhead, as generating structured descriptions requires an MLLM, SDD is superior when semantic relationships, explainability, and generalization across diverse layouts are needed, as in our case.

\begin{table}[!t]
\centering
\small
\renewcommand{\arraystretch}{0.8}
\setlength{\tabcolsep}{4pt}
\resizebox{\columnwidth}{!}{\begin{tabular}{@{}l | c c c @{}}
\toprule
\multirow{1}{*}{\bf In-context Sampling Method} & \multicolumn{1}{c }{\bf GDE (Avg.)} & \multicolumn{1}{c }{\bf Infographic (Acc)} & \multicolumn{1}{c }{\bf Afixa (Acc)} \\
\midrule

Random Selection (9) & 0.735 & 59.88 & 69.15\\
Global Features (6) & 0.740 & 60.05 & 70.23\\
Description-based (6) & 0.745 & 63.21 & 71.44\\
\midrule
\rowcolor{Light}
GRAD (w/o Global Features) (5) & 0.751 & 62.85 & 71.68\\
\rowcolor{Light}
\bf GRAD (4) & \bf 0.757 & \bf 65.97 & \bf 72.17\\

\bottomrule
\end{tabular}}
\caption{\textbf{Ablation on different in-context design selection methods} for \ourapproach$_{\text{GPT-4o}}$. Best results are obtained with GRAD, preserving semantic, spatial and structural information. Best $K$ for top-$K$ retrieval are provided in brackets beside method.}\label{tab:ablation_in_context_gemini}
% \vspace{-2mm}
\end{table}

\begin{table}[!t]
\centering
\small
\setlength{\tabcolsep}{4pt}
\resizebox{\columnwidth}{!}{\begin{tabular}{@{}c | c c | c | c @{}}
\toprule
\multirow{2}{*}{\bf GRAD} & \multicolumn{2}{c |}{\bf SDD} & \multirow{2}{*}{\bf Infographic (Acc)} & \multirow{2}{*}{\bf \ourdataset (Acc)} \\
\cmidrule{2-3}
& w/ Bnd. Box & w/o Bnd. Box & & \\
\midrule

%\cmidrule{2-16}

%F-$1$ & F-$1$ & F-$1$ & F-$1$ & F-$1$ \\

\textcolor{OrangeRed}{\ding{55}} & \textcolor{OrangeRed}{\ding{55}} & \textcolor{OrangeRed}{\ding{55}} & 60.28 & 68.32 \\

\textcolor{OrangeRed}{\ding{55}} & \textcolor{ForestGreen}{\ding{51}} & \textcolor{OrangeRed}{\ding{55}} & 61.13 & 69.85 \\ 

\textcolor{OrangeRed}{\ding{55}} & \textcolor{OrangeRed}{\ding{55}} & \textcolor{ForestGreen}{\ding{51}}  & 61.96 & 71.54 \\

\textcolor{ForestGreen}{\ding{51}} & \textcolor{OrangeRed}{\ding{55}} & \textcolor{OrangeRed}{\ding{55}} & 62.79 & 72.16 \\

\textcolor{ForestGreen}{\ding{51}} & \textcolor{ForestGreen}{\ding{51}} & \textcolor{OrangeRed}{\ding{55}} & 63.71 & 73.29 \\

\rowcolor{Light}
\textcolor{ForestGreen}{\ding{51}} & \textcolor{ForestGreen}{\ding{51}} & \textcolor{ForestGreen}{\ding{51}} & \bf 65.97 & \bf 75.43 \\

\bottomrule
\end{tabular}}
\caption{\textbf{Effect of different components} of \ourapproach$_{\text{Gemini}}$. Results on dataset with layout metadata information are reported. Substantial improvements are found upon adding GRAD and SDD.}\label{tab:ablation_components_gemini}
\end{table}

\begin{table}[!t]
    \centering
    % % \vspace{-15mm}
    \renewcommand{\arraystretch}{1.0}
    \setlength{\tabcolsep}{4pt}
    \resizebox{0.6\linewidth}{!}{\begin{tabular}{@{}c | c c@{}}
    \toprule
    \multirow{1}{*}{\bf $\alpha$} & \multicolumn{1}{c }{\bf Infographics (Acc)} & \multicolumn{1}{c }{\bf Afixa (Acc)} \\
    \midrule
    
    0  & 64.49 & 70.39 \\
    1 & 61.74 & 69.14 \\
    0.25 & 64.96 & 72.03 \\
    0.75 & 63.85 & 70.32 \\
    \rowcolor{Light}
    \bf 0.5 & \bf 65.97 & \bf 72.17 \\
    
    \bottomrule
    \end{tabular}}
        % \vspace{-3.0mm}
        \captionof{table}{\textbf{Ablation on $\alpha$} for \ourapproach$_{\text{Gemini}}$ on Infographic and Afixa datasets.}
    \label{tab:ablation_alpha_gemini}
\end{table}

\begin{table}[!t]
\centering
\small
\renewcommand{\arraystretch}{0.8}
\setlength{\tabcolsep}{4pt}
\resizebox{\columnwidth}{!}{\begin{tabular}{@{}l | c c c @{}}
\toprule
\multirow{1}{*}{\bf Layout Metadata Input Form} & \multicolumn{1}{c }{\bf Infographic (Acc)} & \multicolumn{1}{c }{\bf \ourdataset (Acc)} \\
\midrule

\multicolumn{3}{c}{\ourapproach$_{\text{GPT-4o}}$}\\

\midrule

XML & 68.25 & 75.16\\
\rowcolor{Light}
\bf SDD (Ours) & \bf 69.53 & \bf 76.78\\

\midrule

\multicolumn{3}{c}{\ourapproach$_{\text{Gemini-1.5-Pro}}$}\\

\midrule

XML & 64.26 & 74.98\\
\rowcolor{Light}
\bf SDD (Ours) & \bf 65.97 & \bf 75.43\\

\bottomrule
\end{tabular}}
\caption{\textbf{SDD vs directly passing XML} for \ourapproach$_{\text{GPT-4o}}$ and \ourapproach$_{\text{Gemini-1.5-Pro}}$. Best results are obtained with SDD.}\label{tab:ablation_sdd_xml}
% \vspace{-2mm}
\end{table}

\section{More Qualitative Examples}

\begin{figure*}
    \centering
    \includegraphics[width=0.99\linewidth]{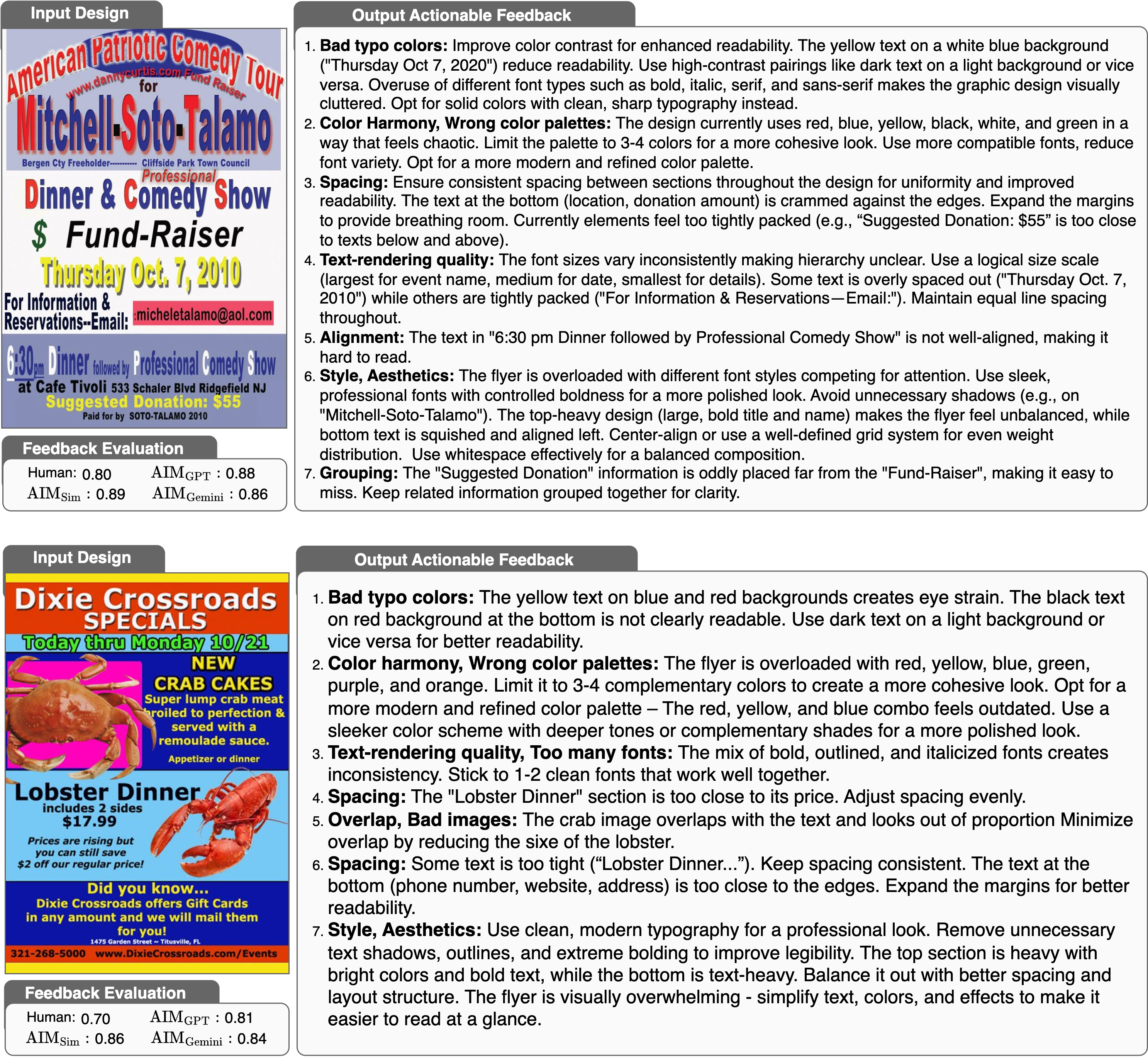}
    \caption{\textbf{Generated feedback along with the design attributes.}}
    \label{fig:qual_1}
\end{figure*}

\begin{figure*}
    \centering
    \includegraphics[width=0.99\linewidth]{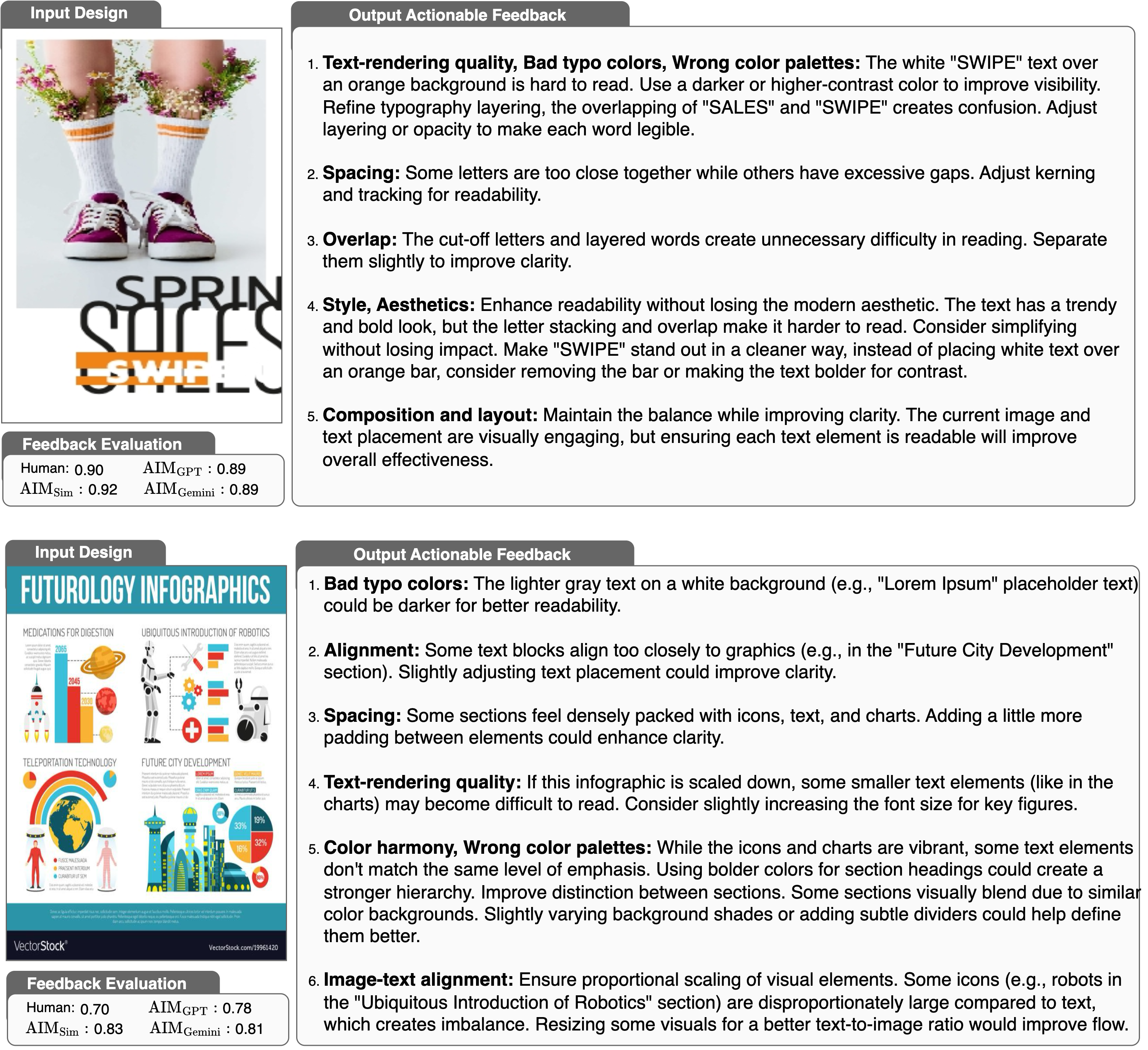}
    \caption{\textbf{Generated feedback along with the design attributes.}}
    \label{fig:qual_2}
\end{figure*}

We provide further qualitative examples in Figs \ref{fig:qual_1} - \ref{fig:qual_2}. 

Additionally, in Figs. 3 - 4, we show that \ourapproach is better than single agent and closer to human evaluation. In Fig. 3, we see the bottom right image where people are 'rafting' instead of 'hiking'. \ourapproach can identify this problem which was possible because we have different experts focusing on specific facets of the design. Whereas a single agent gets distracted with the textual inconsistencies and does not focus on this aspect of semantic alignment between the elements.

Similar observations can be made in the case of Fig. 4, where the single agent is unable to understand the disconnect among the different images, the problems with visual balance and aesthetics, and the grouping of different elements, as reflected from their respective feedback which are validated with the normalized human ratings.

\begin{table*}[!t]
\begin{minipage}[b]{0.31\linewidth}
        \centering
        \begin{subfigure}[b]{\linewidth}
		\includegraphics[width=\linewidth]{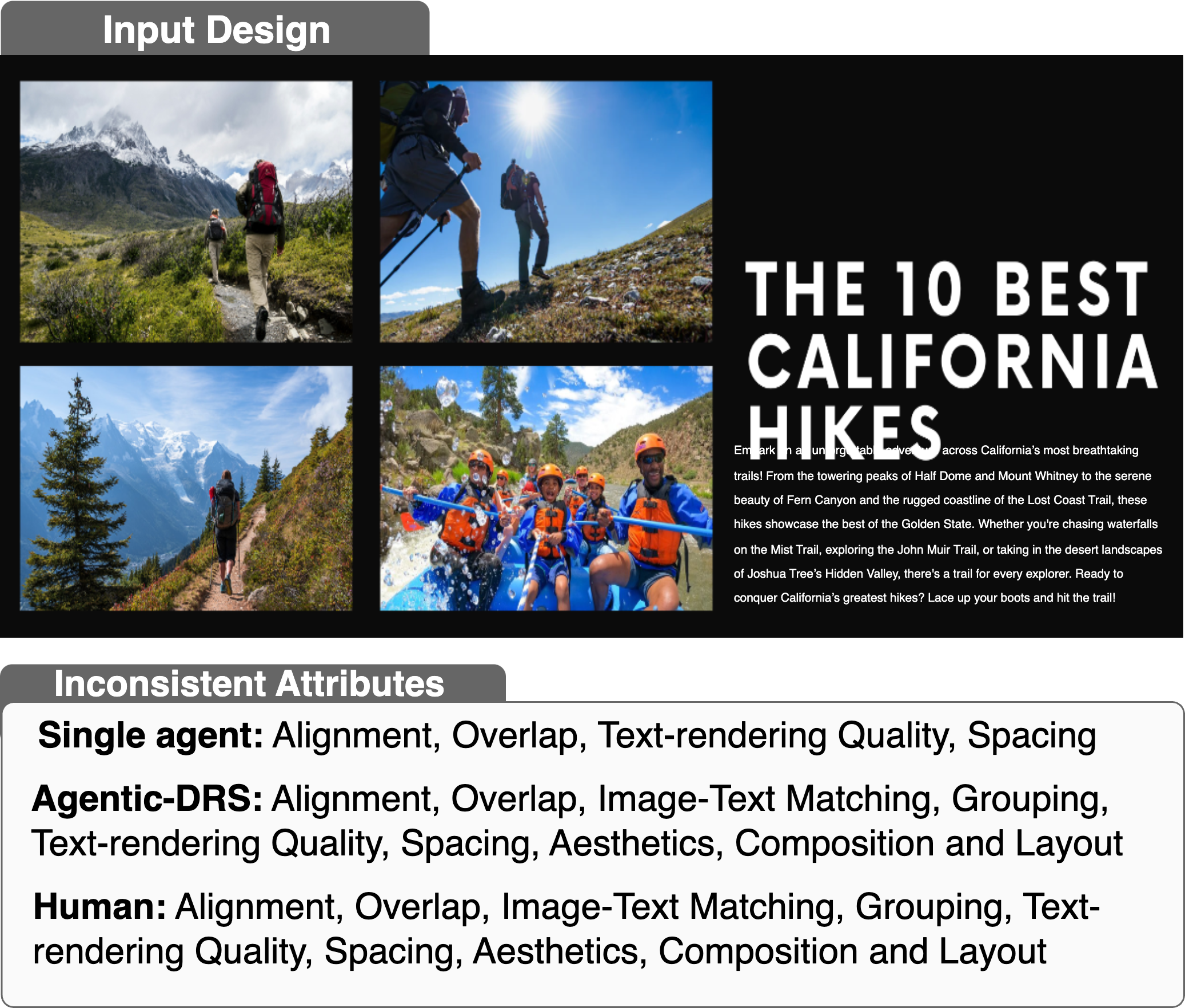}
        \label{fig:qual_3}
        \end{subfigure}
    \captionof{figure}{\textbf{Inconsistent attributes} as identified by \ourapproach, single agent and human raters.}
  \end{minipage}
  \hfill
  \begin{minipage}[b]{0.65\linewidth}
    \centering
        \begin{subfigure}[b]{\linewidth}
		\includegraphics[width=\linewidth]{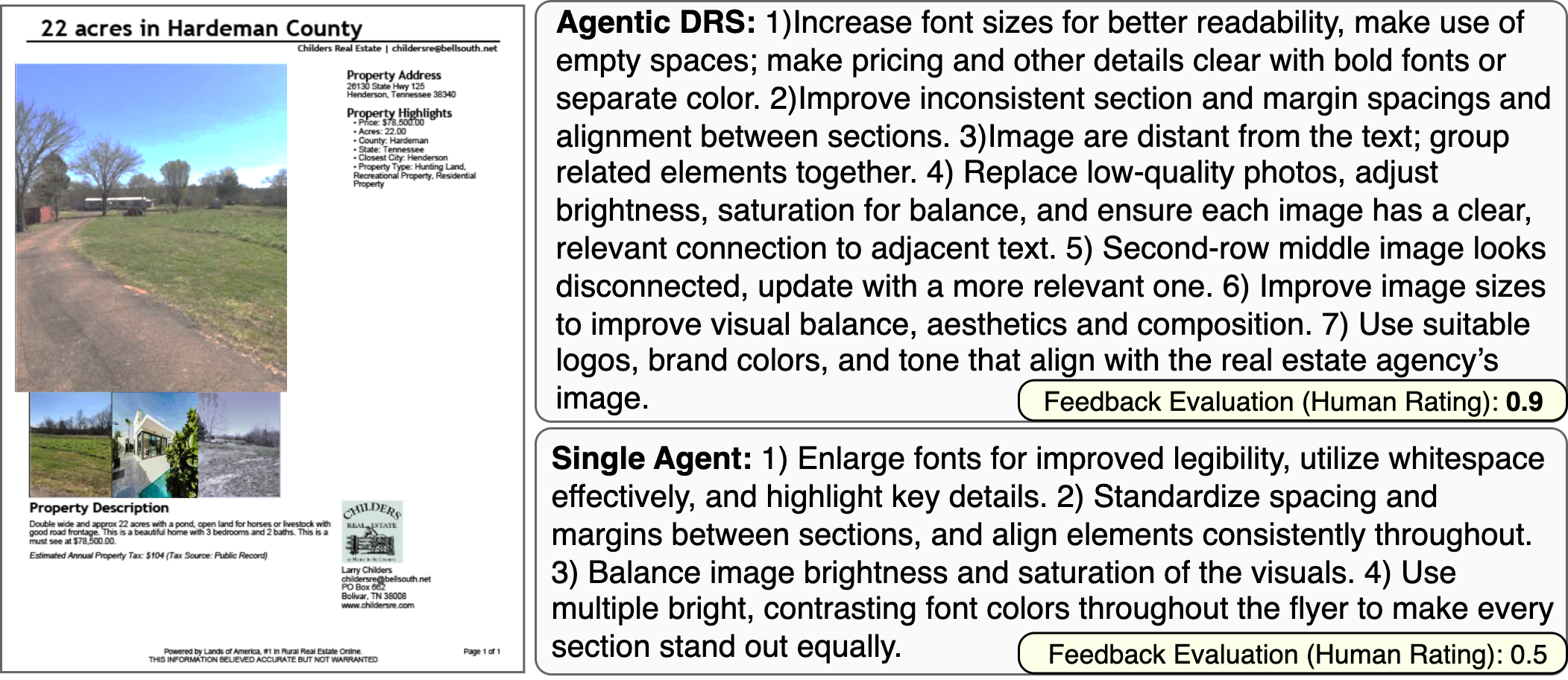}
        \label{fig:qual_4}
        \end{subfigure}
    \captionof{figure}{\textbf{Inconsistent attributes} as identified by \ourapproach, single agent and human raters.}
  \end{minipage}
\end{table*}

\section{Failure Cases}

\begin{figure}[!t]
    \centering
    \includegraphics[width=\linewidth]{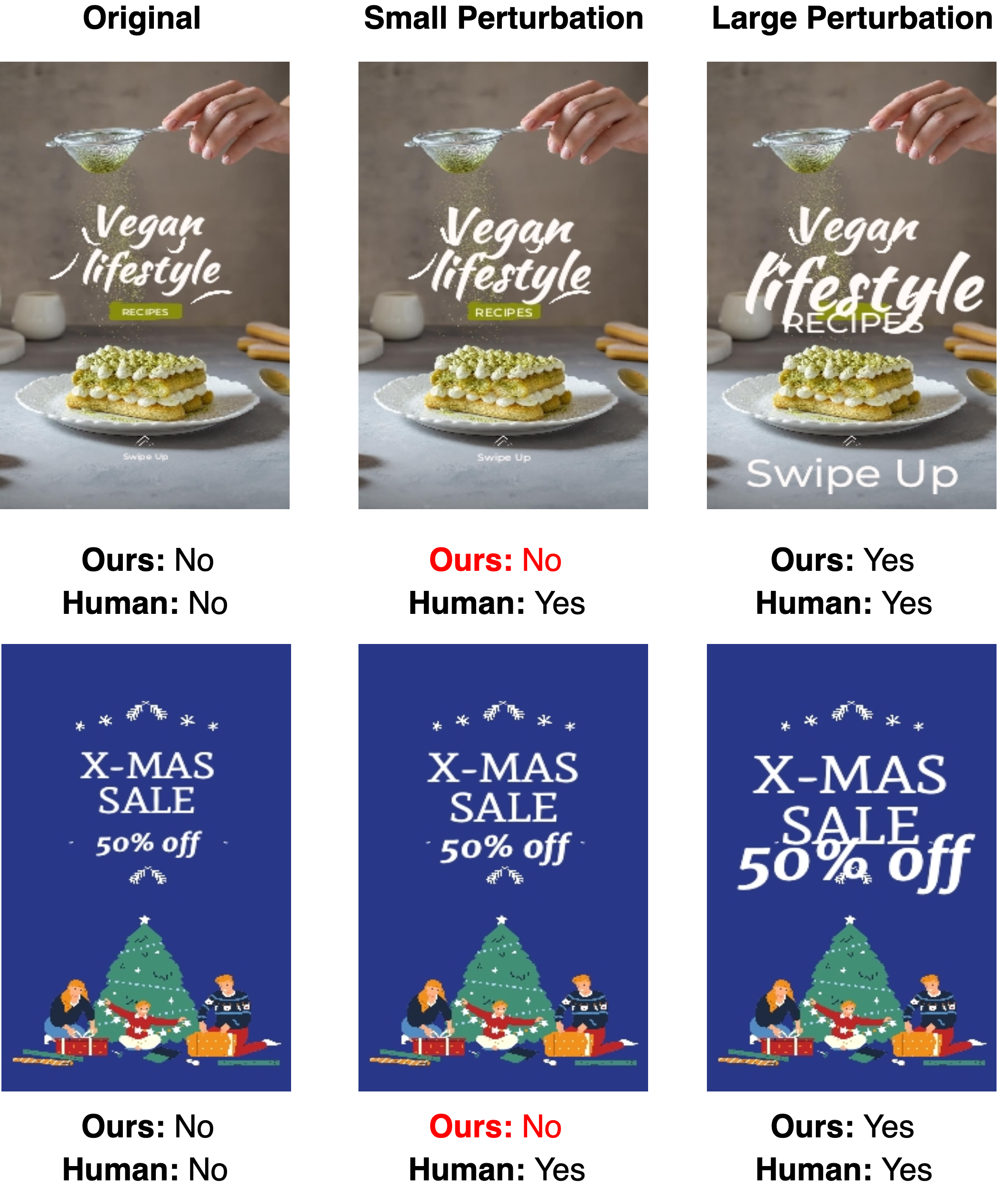}
    \caption{\textbf{Failure Cases of our method.} In the cases of small perturbations our approach fails to recognize the inconsistencies in the design (attribute: `Alignment').}
    \label{fig:failure_cases}
\end{figure}

In Figure \ref{fig:failure_cases}, we show that in the presence of small perturbations, our method fails to recognize the inconsistencies in the design. The result is shown for the attribute `poor alignment elements'. In both the examples in Fig \ref{fig:failure_cases}, the problems in alignment were not well detected as opposed to the human rating where the alignment problem was identified.

\end{document}